\documentclass[final,5p,times,twocolumn,authoryear]{elsarticle}

\makeatletter
\def\ps@pprintTitle{%
 \let\@oddhead\@empty
 \let\@evenhead\@empty
 \def\@oddfoot{}%
 \let\@evenfoot\@oddfoot}
\makeatother

\usepackage{amssymb}
\usepackage{color} 
\usepackage{amsmath}
\usepackage{amssymb}
\usepackage{graphicx}
\usepackage{placeins}
\usepackage{here}
\usepackage{natbib}
\usepackage{multirow}
\usepackage{subfig}

\DeclareMathOperator*{\grad}{\!^\circ}

\begin{document}

\begin{frontmatter}



\title{Pose Estimation and 3D Reconstruction of Vehicles from Stereo-Images Using a Subcategory-Aware Shape Prior}


\author{Max Coenen\textsuperscript{1}, Franz Rottensteiner\textsuperscript{1}}

\address{\textsuperscript{1 }Institute of Photogrammetry and GeoInformation, Leibniz Universit\"at Hannover, Germany - [coenen, rottensteiner]@ipi.uni-hannover.de}

\begin{abstract}
The 3D reconstruction of objects is a prerequisite for many highly relevant applications of computer vision such as mobile robotics or autonomous driving.
To deal with the inverse problem of reconstructing 3D objects from their 2D projections, a common strategy is to incorporate prior object knowledge into the reconstruction approach by establishing a 3D model and aligning it to the 2D image plane. 
However, current approaches are limited due to inadequate shape priors and the insufficiency of the derived image observations for a reliable alignment with the 3D model. 
The goal of this paper is to show how 3D object reconstruction can profit from a more sophisticated shape prior and from a combined incorporation of different observation types inferred from the images.
We introduce a subcategory-aware deformable vehicle model that makes use of a prediction of the vehicle type for a more appropriate regularisation of the vehicle shape.
A multi-branch CNN is presented to derive predictions of the vehicle type and orientation. This information is also introduced as prior information for model fitting.
Furthermore, the CNN extracts vehicle keypoints and wireframes, which are well-suited for model-to-image association and model fitting.
The task of pose estimation and reconstruction is addressed by a versatile probabilistic model.
Extensive experiments are conducted using two challenging real-world data sets on both of which the benefit of the developed shape prior can be shown. 
A comparison to state-of-the-art methods for vehicle pose estimation shows that the
proposed approach performs on par or better, confirming the suitability of the developed shape prior and probabilistic model for vehicle reconstruction.
\end{abstract}




\end{frontmatter}


\section{Introduction}

\begin{figure}[ht]
\centering
		\includegraphics[width=0.999\columnwidth]{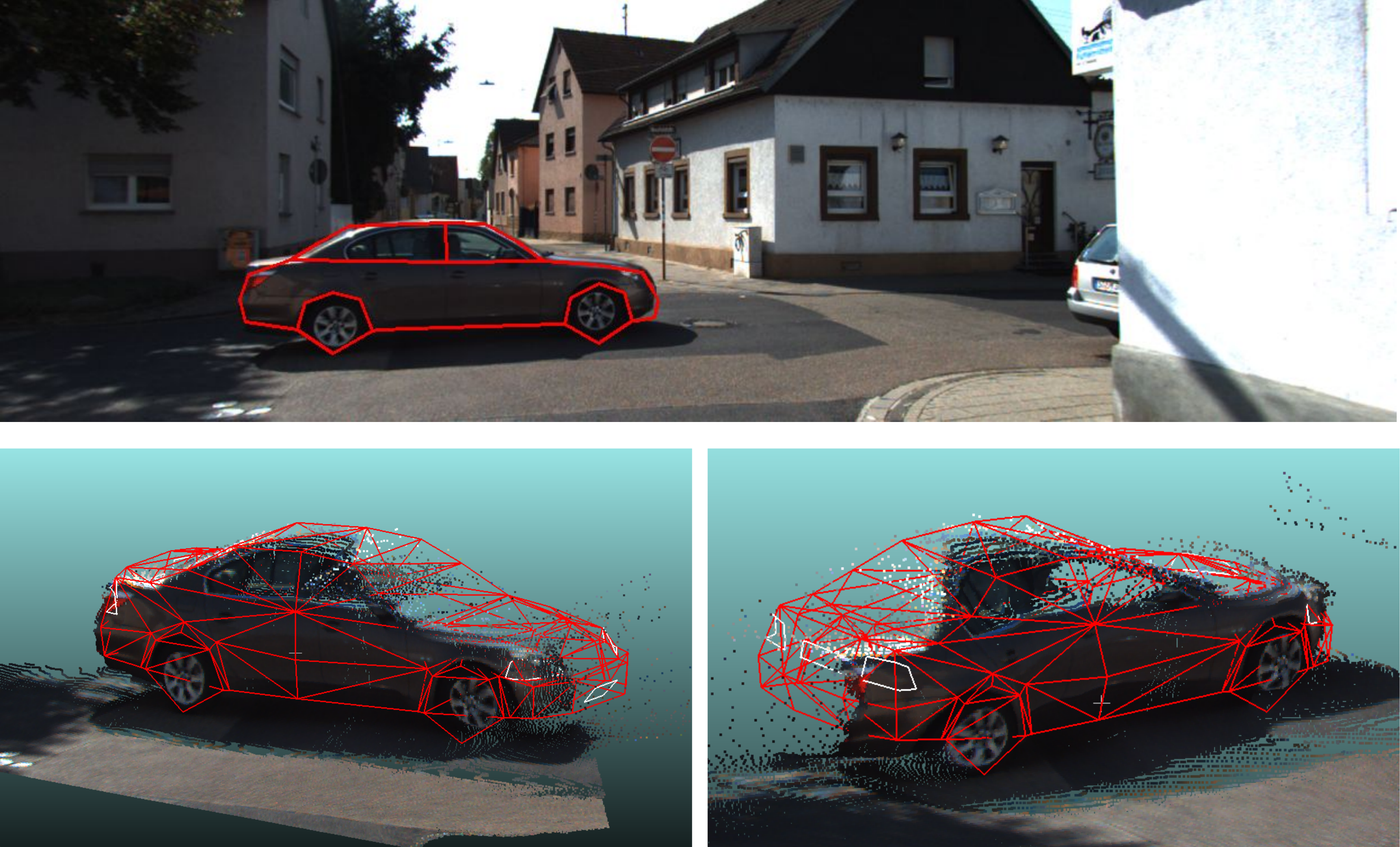}
	\caption{Qualitative result of our method. Top: The 3D wireframe of the fitted vehicle model backprojected to the input image. Bottom: Two 3D views on the reconstructed vehicle model and the stereo points. }
\label{fig:QualResult}
\end{figure}

The image based reconstruction of three-dimensional (3D) scenes and objects is a major topic of interest in computer vision and photogrammetry.
The task of inferring the 3D geometry of objects is very challenging for vision algorithms since the perspective projection from 3D to the 2D image plane leaves many ambiguities about 3D objects, causing their reconstruction and the retrieval of their pose and shape to be ill-posed and difficult to solve.
Nonetheless, 3D scene understanding and 3D reconstruction of specific target objects have a great relevance for several disciplines, e.g. autonomous driving.   
The precise reconstruction of objects, especially of other cars, is fundamental to ensure safe navigation and to enable applications such as interactive motion planning and collaborative positioning. 
Given this background, this paper presents a method for precise vehicle reconstruction from street-level stereo images (cf. Fig.~\ref{fig:QualResult}).
%
%
Geometrically, the image-based reconstruction of 3D objects from their 2D projections is an inverse problem, suffering from the ambiguous mapping inherited by the perspective projection, causing the task to be ill-posed. 
To address the class of ill-posed problems, a common approach is to introduce suitable constraints, e.g.\ derived from prior knowledge about the 3D shape.
Thus, an essential strategy in existing  work on image based object reconstruction is to establish a 3D model and align it to features in the 2D image plane.
To relax the requirement for precisely known object models, parametrised deformable shape priors can be formulated, increasing the set of free parameters by the parameters defining the shape.
To constrain shape deformations, restricting the shape prior to only result in geometrically valid shapes, the common approach is to penalise deviations from the mean shape determined from the training instances, e.g.\ \citep{Zia2,Engelmann}.
However, this procedure is founded on the assumption that the differences of vehicle shapes follow a unimodal distribution centred at the mean shape.
Arguably, this assumption does not hold true and systematically impairs vehicle categories which differ from the average shape.

For object reconstruction usually entities such as keypoints \citep{Pavlakos2017,KrishnaMurthy2017a,Zia2}, edges/contours \citep{Leotta2009,Ramnath2014,Cayon2016}, the surface of the deformable 3D model \citep{Engelmann}, or a combination of all \citep{Coenen2019} are aligned to their corresponding counterparts localised in the image.
Often, the alignment is performed by minimising the reprojection error between projections of the 3D model entities and the associated 2D image detections \citep{Pavlakos2017,Leotta2009}. 
Formulating this goal as an objective function, the estimation of the target
parameters corresponds to the minimisation of that function.
Typically, the objective is non-linear w.r.t.\ the target parameters and typically non-convex, such that one common way for optimisation is to use local optimisation techniques by applying first-order approaches that are based on the
Jacobian of the objective.
For example, \citet{KrishnaMurthy2017} apply an iterative least squares method for the minimisation of the backprojection error of vehicle model keypoints and keypoints detected in the image. 
In \citep{Engelmann, Pavlakos2017}, gradient descent is applied for vehicle model fitting.
However, as non-linear least squares and gradient descent are methods for local optimisation, they do not guarantee finding the global optimum in case the objective function is non-convex.
Instead, convergence to the correct minimum heavily relies on the initialisation of the parameters and consequently, already fairly accurate initial solutions are
required.
Furthermore, least squares optimisation in general is highly sensitive to outliers and erroneous detections which are likely to occur in the context of keypoint based model fitting due to imprecise or false positive keypoint detections.

Following the path of model based object reconstruction, the aforementioned limitations are addressed in this work.
Given initially detected vehicles, we present a methodology for the 3D reconstruction of vehicles from stereoscopic images to finally obtain precise estimates for the vehicle's pose and shape.
In extension to our previous work \citep{Coenen2019}, we make the following contributions in this paper:
\begin{itemize}
\item We propose a new subcategory-aware deformable vehicle model to be used as shape prior. In contrast to existing approaches, e.g. \citep{Zia2,Engelmann}, where a deformable shape model is always learned for the entire class \textit{vehicle}, this work presents a shape model which also considers  individual modes for different vehicle subcategories. 
Thus, the proposed model allows a more detailed shape regularisation if a prediction of the vehicle type is available. The presented shape prior leads to better constraints on the vehicle shape and evidentially also enhances the results of pose estimation.
\item We extend our multi-branch CNN presented in \citep{Coenen2019} to predict a probability distribution for the vehicle's subcategory in addition to the prediction of the vehicle's viewpoint and probability maps for vehicle keypoints and wireframe edges. 
These subcategory predictions are used to apply a more detailed shape regularisation of the vehicles using the proposed subcategory-aware vehicle model.
\item We present a comprehensive probabilistic model for vehicle reconstruction combining multiple observation likelihoods based on the keypoint and wireframe probability maps. Extending our previous work \citep{Coenen2019}, the proposed subcategory-aware shape model and the predictions of the vehicle type are incorporated into the probabilistic formulation to act as a novel state prior for the vehicle shape.
\item We conduct extensive experiments, assessing the effect of individual constituents of the proposed probabilistic model on the quality of the vehicle reconstructions based on the well-known KITTI dataset \citep{KITTI} and our own stereo dataset for vehicle reconstruction. 
Extending the work in \citep{Coenen2019}, we do not only analyse the effects of the proposed shape prior on the pose estimates, but also those of other components of the probabilistic model, and we also evaluate the estimated vehicle shapes.
\end{itemize}
\section{Related work}
This section provides an overview of related work on object reconstruction and object pose and shape recovery.
The focus is on different shape representations used in the literature and on methods that deal with objects, data and applications related to autonomous driving with vehicles as main objects of  interest.

\subsection{3D Shape representations and shape priors}
The ways in which object shapes are represented and prior model knowledge is used vary in the literature. 
In this review, we limit ourselves to the review of related methods for the representation of rigid objects such as vehicles.  
Shape representations which are primarily designed for and applied to non-rigid objects such as humans or clothes, e.g. \citep{Akhter2011}, are not considered.

A strongly generalising and frequently used representation of objects is given by 3D bounding boxes \citep{3DOP,Chen2016,Mousavian2017,Ku2019}. 
The estimation of oriented 3D bounding boxes for the objects implicitly contains the information about the objects' extents and poses, i.e.\ their positions and orientations in 3D space. 
However, a 3D bounding box representation entirely neglects the reasoning about the object's shape by only representing objects as 3D boxes. Yet, the fine-grained estimation of an object's shape can be important for several applications and tasks, such as the (re-)identification of objects \citep{Tang2019}.
To introduce prior model knowledge for the reconstruction of vehicles, simple constraints on the vehicles' symmetry, about the centres of the
wheels or the corners of the rooftop to be in a plane, and prior knowledge about the expected size are enforced in \citep{KrishnaMurthy2017,Wenhao2018}. 
While these constraints give a rather basic and coarse representation of an object, computer-aided design (CAD) models allow a very detailed description of a 3D shape. A large variety of CAD models are available for all kind of different vehicle types and brands which can be used directly to guide the vehicle reconstruction \citep{Displets}. 
However, without knowing the exact CAD model of interest, it is intractable to run computations with each CAD model to find the most suitable one. 
Given a detected vehicle, \citet{DeepManta} use a convolutional neural network (CNN) to predict the closest 3D shape template from a set of rigid reference models and pick this model for further processing. However, potential errors in predicting the shape template may lead to erroneous reconstruction results.
In this work, instead of enforcing a fixed shape for model fitting, softer constraints on the vehicle shape are implemented, which allow deformations of the shape according to the observations. 
For the task of vehicle re-identification, \citet{Tang2019} predict the vehicle type, e.g.\ \textit{compact car, sedan, limousine}, etc., using a CNN. 
However, the predicted type is not utilised in the context of vehicle reconstruction.  
In this work, a CNN which predicts the vehicle type is also used.
A confidence-aware shape prior is presented which makes use of the type predictions by constraining shape deformations during the model fitting according to the predicted confidence scores for different vehicle types. 

In contrast to rigid model instances, deformable shape representations learned from a set of reference shapes are more flexible and allow to cope with the large intra-class variability of vehicles.
Therefore, they are used frequently for object reconstruction. As a consequence,  the shape  parameters are added to the list of target parameters for the estimation.  
In \citep{Engelmann,3DRCNN,Manhardt2019}, a Truncated Signed Distance Function (TSDF) is learned as a deformable shape prior for vehicles from a set of CAD models. 
Using a TSDF, the shape is represented in a voxel grid in which each voxel contains the truncated signed distance towards the object surface and thus, the shape manifold is implicitly represented as the zero-level of the TSDF.
The common shape basis of the training set is learned by applying Principle Component Analysis (PCA) to the TSDF representations of the training samples.
Once the shape basis is learned, any deformed TSDF shape can be encoded by a low-dimensional shape vector. 
Similarly, \citet{DOPS} use an implicit shape representation as a zero level set as a signed distance field. They train a CNN end-to-end to detect cars in 3D point clouds; the CNN also predicts the signed distance for every 3D point, and based on this information, a mesh representation is constructed. 
No quantitative evaluation of shape reconstruction is given. 
In any case, implicit representations only carry the information about the model surface, while information such as semantic keypoint locations or wireframe edges, which can be significant cues for model fitting, are not explicitly contained.
In contrast to that, an Active Shape Model (ASM) is another deformable shape representation frequently used as shape prior for vehicles \citep{Zia2,Lin,KrishnaMurthy2017,KrishnaMurthy2018} that naturally contains keypoint information as it is learned by performing PCA on keypoints from training models. 
In \citep{AAM}, an Active Appearance Model (AAM) is proposed, in which an ASM based statistical shape model is combined with an additional model representing the texture variations associated to the shape variations.
However, while the proposed AAM is only applied to objects observed from one unique viewpoint (a close-up frontal view of human faces in \citep{AAM}, learning the appearance model for objects from arbitrary viewpoints becomes very complex and therefore intractable.
In this work, an ASM representation is adapted as shape prior. However, we extend its keypoint based representation by defining a triangulated mesh and a wireframe topology based on the keypoints to achieve a joint representation for the vehicles surface, keypoints, and wireframe at the same time \citep{Coenen2019}. These model entities are incorporated in the reconstruction process.

Such deformable shape priors are flexible but they can only be deformed in accordance with the variability contained in the training data.
A common way of regularising the degree of admissible deformations during inference is to penalise deviations from the mean shape \citep{Zia2,Engelmann}. However, this strategy is founded in the assumption that the object shape variability follows a unimodal distribution, which usually does not hold true.
Especially in the case of vehicles, the shape variations result in several disjoint modes corresponding to different vehicle types, rather than following an unimodal distribution \citep{Lin}.
Consequently, the applied regularisation of shape deformations is likely to enforce incorrect model shapes.
In this work, the modes resulting from different vehicle types are learned together with the overall ASM representation. 
A CNN based prediction of the vehicle type from the image is used to guide the shape regularisation based on a newly proposed category-aware formulation of the shape prior.

\subsection{Pose estimation and 3D reconstruction}
\subsubsection{3D pose prediction}
With the emergence of CNNs, the prediction of 3D object pose from (single) images has experienced a huge boost, and work on 3D object bounding box prediction has expanded significantly over the last few years.
One line of work follows the two-step procedure that was already successfully applied by Region Convolutional Network (RCNN) approaches for 2D object detection \citep{FasterRCNN}, by generating 3D object proposals in a first step, which are passed through a CNN to generate the final detections in a second step. 
In this context, \citet{3DOP} make use of stereo image data and the 3D information derived from it to introduce geometric priors, such as object height and point density, and to reason about free-space in order to derive 3D bounding box proposals for street-level objects. 
Their follow-up work \citep{Chen2016} replaces the geometric priors by priors based on scene-context and object shape, which are derived from semantic segmentation and instance segmentation using monocular images. While these methods have shown good results, they are computationally expensive due to the generation and the processing of a large number of object proposals initialised in 3D. 
Stereo images are also used in \citep{Peiliang2019}, where a Stereo RCNN is proposed to simultaneously detect and associate 2D bounding boxes in the left and right images. Furthermore, the authors propose to predict keypoints corresponding to the bottom corners of the 3D bounding box in both stereo images in order to derive the oriented 3D box from them. 
However, in this work, we are not only interested in the 3D object bounding boxes but rather aim at a shape aware reconstruction of the vehicles.

Another line of work builds upon the success of existing work on 2D object detection for the task of 3D bounding box estimation. 
\citet{Ku2019} make use of 2D object detections in the image to infer 3D bounding box proposals by leveraging the relation between the 2D bounding box and an estimated object height. 
However, small errors in the 2D bounding box estimates or the height estimates of the 3D bounding box are likely to cause large errors in the position estimation of the object in 3D space.
In \citep{Xiang2018}, a CNN is trained to localise the object centre in the image and to regress the object's orientation and distance to finally derive the object's 3D pose. However, predicting object distances, i.e.\ the absolute scale from single images is an ill-posed problem and therefore causes ambiguous solutions. 
Given 2D detections delivered by an object detector, \citet{Mousavian2017} propose a CNN to regress the object extents and orientation from single images instead of regressing the 3D translation in object space.
The fact that the perspective projection of the 3D bounding box should fit to the 2D image bounding box is used to infer the absolute translation of the object from the regressed object extents and orientation.
Similar to this, \citet{Tekin2018} and \citet{Grabner2018} propose to train a CNN for the prediction of the 2D image locations of the projected 3D bounding box vertices to estimate the 6DoF pose of objects with known size via spatial resection.
However, these approaches are highly sensitive to errors in the 2D predictions and inaccuracies of the regressed parameters. 
Estimating 3D dimensions or 3D pose from single images is highly ambiguous and therefore causes large average 3D position errors which can be up to several meters \citep{Mousavian2017}.
Besides, the mentioned approaches entirely neglect the reasoning about the object shape by only representing objects as 3D boxes.

\subsubsection{3D pose and shape prediction}
Recent work on object reconstruction delivers the 3D object pose together with a set of parameters defining the shape of an object given a parametric shape representation.
\citet{3DRCNN} learn a ten-dimensional shape representation for vehicles by applying PCA to a training set of voxelised vehicle models.
A RCNN is trained to detect vehicles and to regress the ten-dimensional shape vector in addition to the 3D pose parameters, thus obtaining a complete vehicle reconstruction in 3D space.
Instead of using PCA, in \citep{RuiZhu2017,Manhardt2019}, a 3D convolutional autoencoder is trained from voxelised training shapes to learn a shape parameter vector corresponding to the intermediate representation of the autoencoder in the low dimensional latent space. A network is trained to predict a small number of shape parameters together with the object pose. 
A major drawback of such direct approaches, in which a CNN is trained to regress shape and pose parameters, is the strong dependency on usually large amounts of expensive 3D training data.

\subsubsection{Shape aware reconstruction}
In contrast to that, indirect approaches initially detect and finally reconstruct the objects of interest by fitting a 3D model defined a priori to the 2D image observations to reason about the object pose and shape. 
Another advantage of using 3D models as shape priors is their natural invariance w.r.t.\ the viewpoint. 
For instance, a classifier for viewpoint estimation requires training data for each of the considered viewpoint classes, whose number will be large if a fine-grained viewpoint estimation is required, resulting also in a large demand for training data.
In contrast to that, model driven approaches allow for model hypotheses from any viewpoint to be fitted to the image observations.
A common way to shape aware object reconstruction is to match entities such as keypoints, edges/contours, the surface or the silhouette of the model to the corresponding entities inferred from the image.\\

\textbf{Edge/wireframe based reconstruction}\\
Pioneer work on vehicle model fitting was based on model edge to image edge alignment \citep{Tsin2009, Leotta2009}. The authors defined an appearance representation for a deformable vehicle model based on salient object edges that are likely to generate intensity edges such as occluding contours and part boundaries.
Given sufficient 2D-3D correspondences between image and model edges, the pose and shape of the deformable model is estimated using iterative least squares adjustment.
However, finding the 2D image edge corresponding to a 3D model edge, or more specifically finding corresponding point pairs on these edges, is non-trivial and a challenging task.
Occlusions, illumination conditions, contrast, shadows or reflections, and model initialisation are likely to cause outliers and consequently lead to incorrect correspondences.
An attempt to filter edge maps and extract semantically meaningful contours has been presented in \citep{Isola2014} and is used in \citep{Cayon2016} to reduce the number of outliers for the task of edge-to-edge alignment.
Furthermore, a threshold for the angles between the edge normals of image and model edges is used as a criterion by \citet{Cayon2016} and \citet{Ramnath2014} to discard unlikely correspondence candidates.
Still, the risk of incorrect matches and the need for good model initialisation remains. 
In this paper, a CNN is used to extract the  desired vehicle wireframe edges, where the CNN is trained to semantically distinguish between wireframe edges belonging to different sides of a vehicle, which allows an informed, more robust, and initialisation-invariant model-to-image-edge association.\\

\textbf{Contour/silhouette based reconstruction}\\
Another strategy for model based object reconstruction is to align the silhouette that results from the 3D model to a predicted segmentation mask of the target object. 
\citet{Prisacariu2012} train a Random Forest based on Histogram of oriented gradients (HoG) \citep{HoG2005} and colour features for foreground-background classification and define an energy function considering pixelwise foreground and background matching scores to fit a deformable vehicle model to vehicle detections. 
A similar energy term is incorporated in \citep{Dame}, where foreground-background models are learned from reference segmentations for the parts of a Deformable Part Model (DPM) \citep{DPM} detector, which is used to infer pixel-wise foreground-background probabilities during test time.
In \citep{Kar2015}, an instance segmentation method is applied to derive segmentation masks and a silhouette consistency term is used in the model fitting procedure. 
Similarly, \citet{RuiWang2020} propose  a silhouette alignment term measuring the consistency between the an image segmentation mask and the object mask obtained by projecting the shape embedding into the image and combine it with a term enforcing photometric consistency between the two images of a stereo pair in order to fit a 3D SDF into detected vehicles. 
However, the mapping from a 3D shape model to a 2D image silhouette is highly multi-modal and ambiguous, e.g.\ because object details and structures within the silhouette are neglected, and moreover, segmentation masks of a vehicle in front and in rear views are almost identical, which makes the alignment highly sensitive to initialisation. Furthermore, it is also sensitive to segmentation errors and occlusions.\\

\textbf{Surface based reconstruction}\\
To reconstruct objects based on the 3D surface of an object shape model requires depth or 3D information that can be derived from image observations (e.g.\ from stereo triangulation or Structure from Motion (SfM)) or can be obtained from laserscanning. 
\citet{Displets} use CAD models of vehicles to sample disparity patches from them and align these patches to disparity maps estimated from stereo images.
Similarly to this, in \citep{OSF} a 3D vehicle ASM is also estimated in accordance with a dense disparity map. In addition, images of a subsequent time epoch are used to also incorporate model constraints based on scene flow information. 
In contrast, this paper proposes a method for vehicle reconstruction using a single stereo image pair. 
In \citep{Engelmann} and \citep{Coenen2017}, a TSDF and an ASM, respectively, are learned as shape priors for vehicles and are fitted to 3D point clouds obtained from dense stereo correspondences.
Similarly, \citet{StreetSideDet} fit a vehicle ASM to 3D points obtained from mobile laserscanning. 
Points reconstructed in 3D from multi-view images are used by \citet{Cayon2016} to fit a CAD model to detected vehicles.
In these cases, model fitting is based on minimising the distances between the 3D points and the surface of the shape manifold.
This procedure presents various difficulties.
One of the major problems is the missing semantic information of 3D points, required to relate them to corresponding parts of the vehicle model. 
Another problem arises from outliers in the 3D point cloud, resulting e.g.\ from detection, matching or segmentation errors, and points belonging to parts not represented by the generalised shape prior, like the vehicle's interior, antennae, mirrors, etc.
Furthermore, noisy point clouds and the increasing uncertainty of the stereo reconstructed 3D points with increasing distance from the camera remain challenging for reconstruction approaches that are only based on 3D points.
Besides, when 3D points are used as the only data source, valuable image cues are disregarded completely for model fitting. 
In this work, 3D information is used jointly with 2D image information to exploit synergies of both domains. \\

\textbf{Keypoint based reconstruction}\\
Another strategy for shape-aware vehicle reconstruction is to match model keypoints with their corresponding keypoints detected in the image.
One advantage of using such semantic keypoints compared to the approaches described so far is the easier definition of correspondences between the image and model entities.
Traditionally, handcrafted features, often based on HoG features, are used for the model-to-keypoint alignment, e.g.\ in \citep{Li2011, Zia2, Bao}, while recently CNNs were introduced to detect keypoints, resulting in a better performance.
For instance, \citet{DeepManta} extend a RCNN for object detection to also regress keypoint coordinates in addition to the bounding box and the object class.
A more frequently applied architecture for keypoint detection is a stacked-hourglass architecture \citep{Pavlakos2017, KrishnaMurthy2017a, Wenhao2018}, in which multiple stacked U-Nets \citep{unet} are used to infer keypoint probability maps from which the keypoint locations are derived, e.g., using non-maximum suppression.
A U-Net-like architecture is also used in this work to predict keypoint heatmaps.
Given the 2D image keypoint locations and their corresponding vertices of the 3D model, one naive approach for model fitting is to minimize the reprojection error of a spatial resection to solve for the 6DoF pose \citep{DeepManta}.
However, this naive procedure is problematic for various reasons.
On the one hand, it is intolerant to keypoint localisation errors, while at the same time the inferred 2D keypoint localisations are likely to be imprecise, leading to potentially large errors in 3D.
On the other hand, it is not robust w.r.t to outliers such that potentially false keypoint detections influence the pose estimation directly and demand for robust strategies. 
The probabilistic approach presented in this paper avoids the need for precise keypoint locations, because it is not built upon inferred 2D keypoint coordinates but on the raw keypoint probability maps instead.
By incorporating the full keypoint probability distributions into the optimisation instead of only their inferred modes, the model fitting gains robustness w.r.t\ imprecise keypoint localisations caused e.g.\ by broad probability distributions. 
Performing the keypoint alignment on the raw heatmaps has also been done by \citet{Zia2}. The authors used a Random Forest (RF) classifier \citep{Breiman2001}, trained on gradient based handcrafted features for keypoint classification to derive the keypoint probability maps. The performance of a RF compared to deep learning based techniques, however, is qualitatively inferior as has been found by own previous work using a RF \citep{Coenen2019b} and a CNN \citep{Coenen2019} for the prediction of vehicle keypoints. 

\section{Methodology} \label{sec:Methodology}
This section presents our approach for the model based shape and pose recovery of initially detected vehicles from stereo imagery. 

\subsection{Overview}\label{sec:MethOverview}
The goal of the proposed method is to recover the precise 3D pose (i.e.\ the 6DoF parameters of the position and orientation in 3D) as well as the type and shape of vehicles detected from street-level stereo images. 
The method requires images acquired by a calibrated stereo rig, e.g. attached to a moving platform. 
The camera synchronisation is assumed to be sufficiently accurate so that the influence of object and platform  movements can be neglected. 
The interior and relative orientations as well as the length of the baseline are assumed to be known, and the images are rectified prior to processing so that epipolar lines correspond to image rows. 
The left stereo partner is defined to be the reference image.
The real-time capable ELAS matcher \citep{ELAS} is applied  to determine a dense disparity map for every stereo pair, which  is used to reconstruct a 3D point cloud in the 3D model coordinate system ${^M}C$ from every pixel of the reference image via triangulation.
The origin of the model coordinate system is defined to be the projection centre of the left camera. 
Its ${^M}X/{^M}Y$ plane is parallel to the image plane of the epipolar images and the $^{M}Z$-axis points into the viewing direction.

The proposed method is designed to deduce a 3D vehicle model in ${^M}C$ which represents the detected vehicle best in terms of pose and shape.
To this end, a 3D model is fitted to the observed data.
The target parameters are implicitly contained in the fitted model. 
Prior knowledge about the 3D layout of the observed scene is extracted and used to constrain the parameter space of the model fitting approach. 
A deformable vehicle model is learned as a shape prior and is fitted to the detected vehicles. 
The detection of vehicles and their reconstruction are treated as decoupled tasks.
To initially detect the vehicles visible in the stereo images, a state-of-the-art detection method is adapted and its output is tailored towards the requirements for the proposed vehicle reconstruction method.

\begin{figure}[ht]
\begin{center}
		\includegraphics[width=1.0\columnwidth]{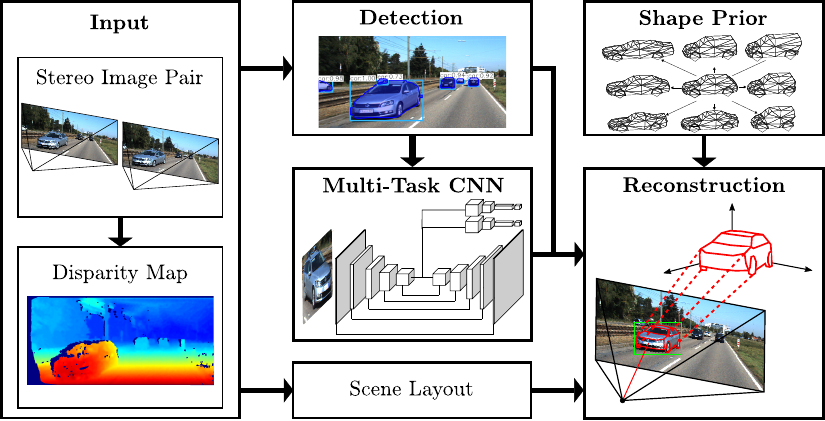}
	\caption{Overview of the proposed framework.}
\label{fig:overview}
\end{center}
\end{figure}

Based on the initial vehicle detections, the core of the proposed method consists of
\textbf{(1)} a novel subcategory-aware deformable vehicle model,
\textbf{(2)} a multi-task CNN, trained to extract various pieces of semantic information from the vehicle detections to be incorporated into 
\textbf{(3)} an extensive probabilistic model that is designed to find the best fitting instance of the deformable model for each vehicle.
The general framework of the proposed approach is depicted in Fig.~\ref{fig:overview} and will be presented in the following sections.

\subsubsection{Problem statement}
The ultimate goal of the proposed method is to detect the set of vehicles $\mathcal{V}$ visible in the given stereo pair and to recover their precise shape and 6DoF poses in the coordinate system ${^M}C$. 
For this purpose, we describe each stereo scene by a 3D ground plane $\Omega \in \mathbb{R}^3$ (cf. Sec..~\ref{sec:ScenePrior}). 
Furthermore, each vehicle $v \in \mathcal{V}$ is associated with its state vector $\mathbf{s}=(\mathbf{t}, \theta, \gamma)$.
The state vector contains the pose and shape of the vehicle represented by its position $\mathbf{t}$ on the ground plane, its orientation $\theta$, i.e.\ the rotation angle about the normal vector of the ground plane, and a vector $\gamma$ determining the shape of a deformable ASM representing the vehicle (cf.\ Sec.~\ref{sec:ShapePrior}).
It has to be noted that the 6DoF vehicle pose parameters w.r.t.\ the reference camera can easily be derived from the 2D pose $\mathbf{t}$ and $\theta$ on the ground plane, knowing the rigid transformation between the ground plane and the model system ${^M}C$.

\subsubsection{Scene layout} \label{sec:ScenePrior}
Prior to vehicle reconstruction, the stereo data are used to derive knowledge about the 3D layout of the scene, represented by the 3D ground plane and a probabilistic free-space grid map. Requiring vehicles always to be located on the ground plane, estimating that plane reduces and constrains the parameter space of the model fitting approach, predetermining three of the 6DoF vehicle pose parameters (1 rotational and 2  translational parameters).
The \textbf{ground plane} $\Omega$ is extracted from the reconstructed 3D points using the RANSAC-based method described in \citep{Coenen2018}. 
All inliers of the final RANSAC consensus set are stored as the set of ground points $\mathbf{X}_\Omega$, whereas
%
the remaining points 
form the set of arbitrary object points $\mathbf{X}_{Obj}$.

A \textbf{probabilistic free-space grid map} $\Phi$ defined in the ground plane with a spatial resolution $l_\Phi$ is created to quantitatively represent areas not occupied by any object \citep{Coenen2018}. 
For each grid cell $\Phi_g$ with $g \in [1,G]$ the number of ground points $n^g_\Omega$ and the number of object points $n^g_{Obj}$ whose orthogonal projection falls in the respective cell are counted. 
Grid cells without any projected points are marked as \textit{unknown}.
For all the other cells, the probability $\rho_g$ of the cell to be free-space is calculated according to
\begin{equation}
\rho_g = \frac{n^g_\Omega}{n^g_\Omega + n^g_{Obj}}.
\end{equation} 
The free-space grid map is used to derive prior information about the vehicle's position (cf.\ Sec.~\ref{sec:ProbModel}). 

\subsubsection{Detection of vehicles} \label{sec:VehicleDetection}
To detect the vehicles $\mathcal{V}$ we apply the pretrained Mask-RCNN of \citet{maskRCNN} to the reference image due to its good performance.
It does not only deliver bounding boxes but also an instance segmentation mask for every vehicle detection $v \in \mathcal{V}$.
Each detection is associated with an observation vector $\mathbf{o}^{det} = (\mathbf{X}_v, {^l}\mathcal{B}, {^r}\mathcal{B}$) containing a set of object points $\mathbf{X}_v$, as well as its bounding boxes ${^l}\mathcal{B}$ and ${^r}\mathcal{B}$ in the left and right images, respectively. 
To extract the vehicle points $\mathbf{X}_v$, the 3D points reconstructed from the disparities of the foreground pixels belonging to the respective segmentation mask are chosen.
While the vehicle bounding boxes in the left image ${^l}\mathcal{B}$ are delivered as an output of the Mask-RCNN, the bounding boxes ${^r}\mathcal{B}$ in the right image are derived from the dense stereo correspondences.

\subsection{Subcategory-aware shape prior} \label{sec:ShapePrior}
Similar to \citet{Zia2} we use a 3D ASM as vehicle shape prior. 
The ASM is learned by applying principal component analysis (PCA) to a set $\mathcal{K}$ of a total number $C_\mathcal{K}$ of 3D keypoints that were manually labelled for a variety of CAD models of vehicles belonging to one of a set of different vehicle types $\mathcal{T}$.
In the experiments of this paper, seven vehicle types are distinguished with $\mathcal{T} = $~\textit{\{Compact\ Car,\ Sedan,\ SUV,\ Estate\ Car,\ Sports\ Car,\ Truck,\ Van\}}.
A synthesised model, deformed according to the shape parameters $\gamma$, is denoted by $M(\gamma)$ and can be obtained by the linear combination
\begin{equation}
M(\gamma) = \mathbf{m} + \sum_{s} \gamma^{(s)} \ \sigma_s \  \mathbf{e}_s \label{eq:ASM}
\end{equation}
of the mean model $\mathbf{m}$ and the eigenvectors $\mathbf{e}_s$, weighted by the square roots of their corresponding eigenvalues $\sigma_s^2$ and scaled by the object specific shape parameters $\gamma^{(s)}$.
It has to be noted that in the practical realisation of the presented method, not all eigenvalues and eigenvectors of the ASM are considered. 
Instead, in order to reduce the dimensionality of the unknown shape parameter vector $\gamma$, the number $n_s$ of considered shape parameters is restricted and is defined as a proper tradeoff between the complexity of the model and the quality of the model approximation.
A fully parametrised instance of a 3D vehicle ASM, denoted by $M(\mathbf{s})$, can be created by shifting and rotating the deformed model $M(\gamma)$ on the ground plane according to the translation vector $\mathbf{t}$ and the heading angle $\theta$.

\subsubsection{Geometrical representation}
\begin{figure*}[ht]
\centering
		\includegraphics[width=0.99\textwidth]{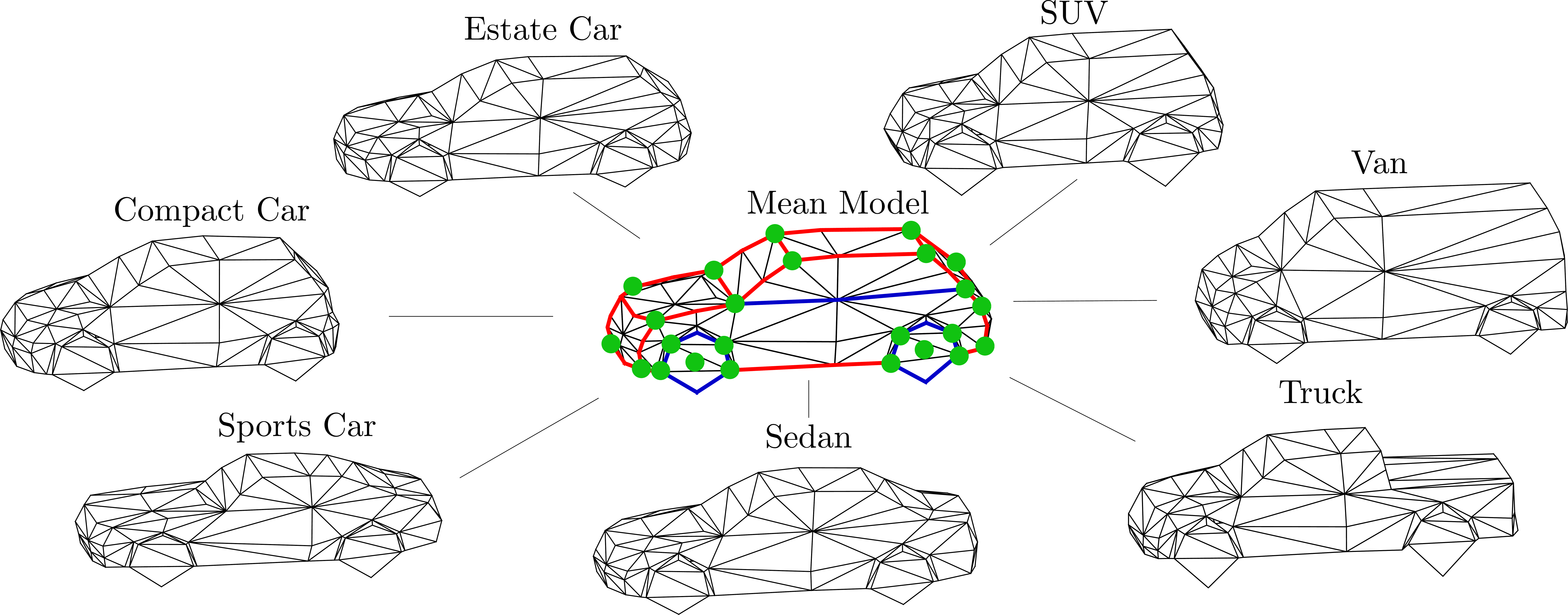}
	\caption{Visualisations of the ASM. In the centre, the mean model is depicted with \textit{crease} edges in red and \textit{semantic} edges in blue. The triangulated surface is shown in black, the \textit{appearance} keypoints in green. Additionally to the mean model, deformed models corresponding to the modes $\gamma^\tau$ for seven vehicle categories are shown.}
\label{fig:ASM}
\end{figure*}

While the classical representation of an ASM only contains explicit information about the keypoints \citep{ASM}, in this work, the ASM is enriched by an additional explicit definition of the model surface as well as a definition of model wireframe edges. 
In particular, a triangular mesh $M_\Delta$ is defined for the ASM vertices $\mathcal{K}$ to represent the model surface.
The topology of the mesh representing the surface of the ASM is manually defined once and kept constant for all generated, deformed models.
A visualisation of the defined triangular vehicle mesh can be found in Fig.~\ref{fig:ASM}. 
Note that keypoints exist (e.g.\ the centre points of the wheels) which are not part of the triangulation in order to decrease the number of faces. 

Moreover, edges connecting pairs of keypoints are manually chosen to define a wireframe topology $M_\mathcal{W}$ of the vehicle model, consisting of both, \textit{crease} edges that describe the outline of the vehicle, and \textit{semantic} edges describing the boundaries between semantically different vehicle parts.
Each entry of $M_\mathcal{W}$ contains a tuple of keypoints from $\mathcal{K}$ defining a single edge of the wireframe.
The edges chosen for this purpose are also depicted in Fig.~\ref{fig:ASM}.
Furthermore, the wireframe $M_\mathcal{W}$ is subdivided into four groups of wireframe edges $M_\mathcal{W}^w$, each of which contains all edges in $M_\mathcal{W}$ belonging to the wireframe of one of the four vehicle sides $w \in \{\text{front, back, left, right}\}$. Note that the edges in the four wireframe subsets are not mutually exclusive, as a wireframe edge can belong to two of the distinguished vehicle sides. This distinctive wireframe representation adds further semantic information to the edges.
The approach for vehicle reconstruction proposed in this work makes use of this semantic wireframe distinction by learning a CNN based detector for each of the wireframe representations $M_\mathcal{W}^w$ (cf.\ Sec.~\ref{sec:CNN}).

In addition to that, a subset of keypoints is chosen to contain the \textit{appearance} keypoints $\mathcal{K}_A \subset \mathcal{K}$ for which an image based detector is learned as well (cf.\ Sec.~\ref{sec:CNN}). This set of keypoints contains a number $C_A$ of keypoints with a potentially distinctive appearance, such as centre points of wheels, corner points of the wind shield and the rear window, front and back lights, etc.
In comparison to the commonly used keypoint based ASM representation \citep{Zia2,Pavlakos2017}, the ASM proposed in this paper is extended by the explicit definition of the 3D surface, by the wireframe definition described earlier, as well as by the subcategory awareness which is described subsequently in Sec.~\ref{sec:ModeLearning}. 
The triangulated surface as well as the wireframe and \textit{appearance} keypoint definition can be seen in Fig.~\ref{fig:ASM}.
%
%
%
\subsubsection{Mode Learning} \label{sec:ModeLearning}
In the literature, a common practice  is to constrain the shape parameters according to the deviations from the mean shape $\mathbf{m}$ \citep{Zia2,Engelmann}, i.e.\ according to the deviations of $\gamma$ from the zero vector (cf.\ Eq.~\ref{eq:ASM}).
This strategy is based on the implicit simplifying assumption of a unimodal distribution of the vehicle shape deformations represented by the underlying principal components of the ASM, but in reality the distribution is expected to have several disjoint modes for different vehicle classes.
For instance, let $\mathcal{T}$ only contain the types \textit{Van} and \textit{Sports Car}; in this example, the mean model represents a shape which is neither a \textit{Van} nor a \textit{Sports Car}, and penalising the deviations from the mean shape is  not an optimal choice.
To obtain a more realistic model, the mode for each of the vehicle types in $\mathcal{T}$ is determined in addition to the general ASM formulation. 
During reconstruction, a prediction of the vehicle type (cf.\ Sec~\ref{sec:CNN}) is used to penalise deviations from the corresponding mode instead of the global mean shape.
We denote the mode for every vehicle type $\tau \in \mathcal{T}$ by $\gamma^\tau$, a vector representing the shape parameters that describe the mean shape $\mathbf{m}^\tau$ of all training exemplars associated to the respective type $\tau$. 
Following Eq.~\ref{eq:ASM}, the mean category shape and its residuals $\mathbf{v}_\tau$ are expressed by 
\begin{equation}
\mathbf{m}^\tau + \mathbf{v}_\tau = \mathbf{m} + \sum_{s=1}^{n_s} \gamma_s^\tau \sigma_s \mathbf{e}_s.
\end{equation}
The modes are calculated according to a least-square fitting of the overall ASM to the mean shape of each type by
\begin{equation}
\gamma^\tau = (\textbf{A}^\text{T} \textbf{A})^{-1} \textbf{A}^\text{T} (\mathbf{m}^\tau - \mathbf{m}), \label{eq:ASMModeLearning}
\end{equation}
where $\mathbf{m}$ is the mean model of all training exemplars. The mean shape of each type $\mathbf{m}^\tau$ is calculated in advance, given the CAD models and their annotated vehicle categories. In this context, it has to be noted that the only additional annotation effort that is required for the proposed mode learning approach consists of associating a vehicle type label to each CAD model used for learning the ASM. 
The Jacobi matrix $\mathbf{A}$ is calculated from the partial derivatives of the linear Eq.~\ref{eq:ASM} by the shape parameters $\gamma_s$. 
With $C_\mathcal{K}$ keypoints defining the ASM and $n_s$ parameters that are considered during vehicle reconstruction, the dimensions of the Jacobi matrix are equal to $[C_\mathcal{K} \cdot 3 \times n_s]$. 
The mean ASM model as well as the deformed models according to the modes $\gamma^\tau$ of seven categories are shown in Fig.~\ref{fig:ASM}.

\subsection{Multi-branch CNN} \label{sec:CNN}
\begin{figure*}[ht]
\centering
		\includegraphics[width=0.99\textwidth]{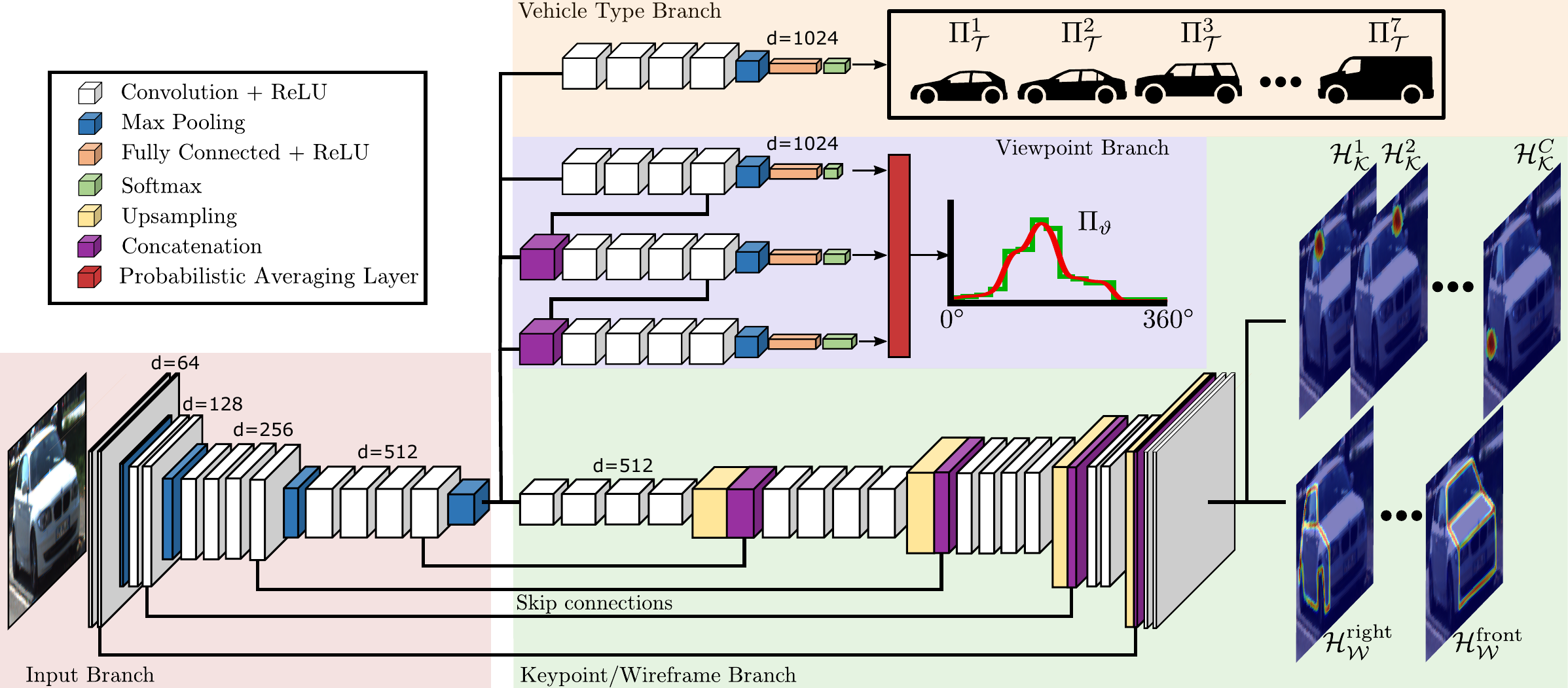}
	\caption{Architecture of the multi-branch CNN. The input is a 3 channel image of size 224x224. The convolutional filters have size 3x3, max pooling and upsampling use filter size 2x2 and stride 2. The number of filters is denoted by d in the figure. Further explanations are given in the main text.}
\label{fig:VehicleCNN}
\end{figure*}

As a first step of the model-based vehicle reconstruction, a mulit-branch CNN is trained and used to infer semantic information to be used in the probabilistic model.
We use the same multi-branch CNN as proposed in \citep{Coenen2019}, but add an additional branch for the prediction of the vehicle type.
To make this paper self-contained, we will describe the complete CNN architecture in this section.
The input to the network is an image showing a vehicle, cropped by the detection bounding box. 
The multi-branch CNN consists of one common input branch and three individual output branches, each of them corresponding to one task. Unless specified differently, 3x3 filters are used in the convolutional layers and 2x2 filters with stride 2 are used for max pooling and upsamling operations. 
The overall architecture of the network can be seen in Fig.~\ref{fig:VehicleCNN}.
All vehicle detections in the reference (left) image, cropped by their respective bounding box, are fed into the network to infer a probability distribution $\Pi_\mathcal{T}$ for the vehicle's type, a probability distribution $\Pi_{\vartheta}$ for the vehicle's viewpoint, and the keypoint and wireframe probability maps ${^l}\mathcal{H}_\mathcal{K}$ and ${^l}\mathcal{H}_\mathcal{W}$, respectively.
Additionally, the keypoint and wireframe heatmaps ${^r}\mathcal{H}_\mathcal{K}$ and ${^r}\mathcal{H}_\mathcal{W}$ are computed for the corresponding bounding box crops of the right image by feeding the detection window through the keypoint/wireframe branch.
The results are gathered in an additional observation vector $\mathbf{o}^{cnn} =({^l}\mathcal{H}_\mathcal{K}, {^l}\mathcal{H}_\mathcal{W}, {^r}\mathcal{H}_\mathcal{K}, {^r}\mathcal{H}_\mathcal{W})$ for every detected vehicle. 
%
%
%
\subsubsection{Input branch} \label{sec:CNNInputBranch}
The input to the network is a 3-channel image of size 224x224. Bilinear interpolation is used to resize image crops of the vehicle detections to the required size. 
The input branch acts as a shared backbone feature extractor of the CNN.
The architecture of the VGG19 network \citep{VGG19} is adopted for this purpose. 
The feature map of size 14x14 produced by the \textit{input branch} is forwarded to the task specific branches which are explained in the following sections.

\subsubsection{Viewpoint branch} \label{sec:ViewpointBranch}
\begin{figure}[ht]
\centering
\subfloat[] {\includegraphics[width=0.45\columnwidth]{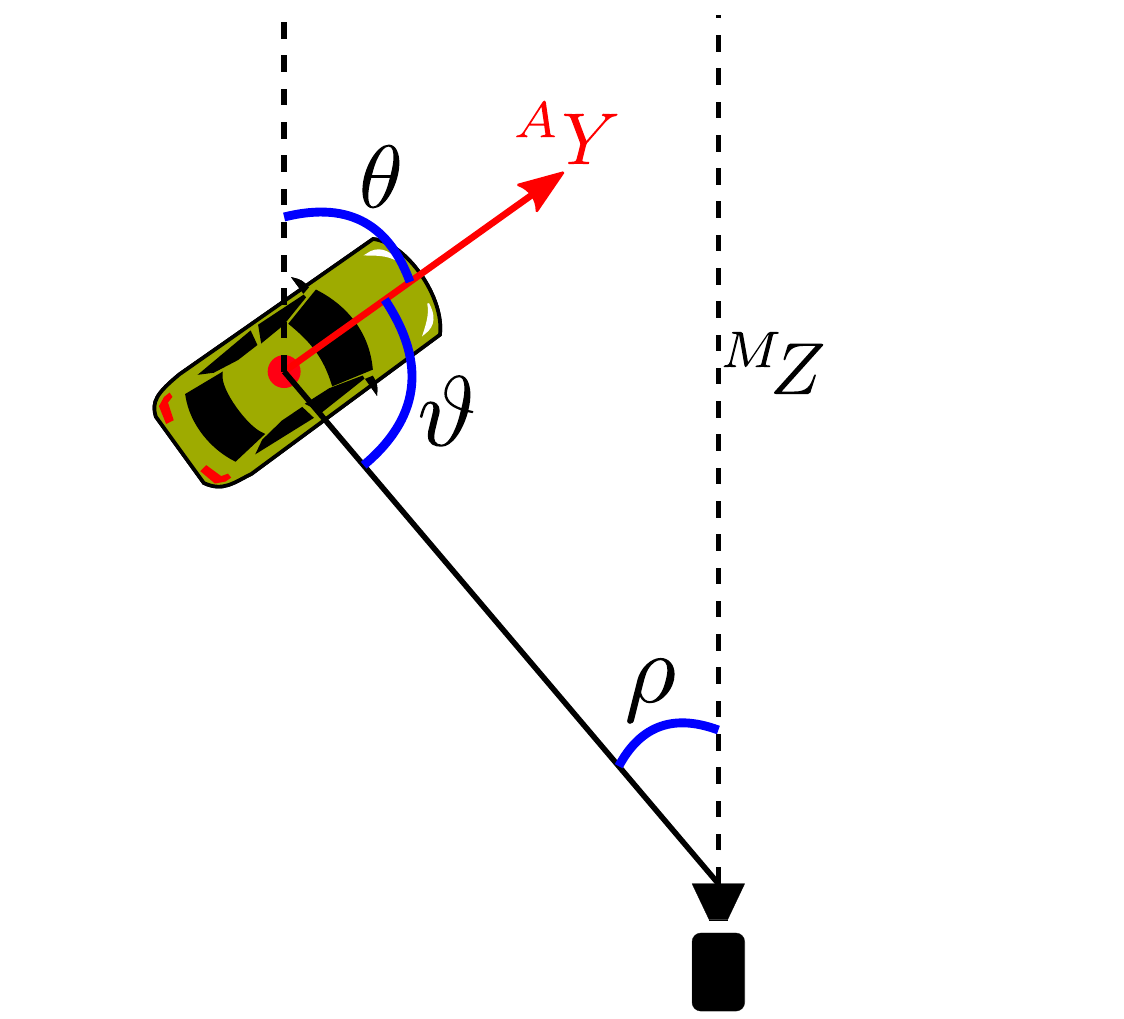}\label{fig:ViewpointDef}} 
\subfloat[] {\includegraphics[width=0.45\columnwidth]{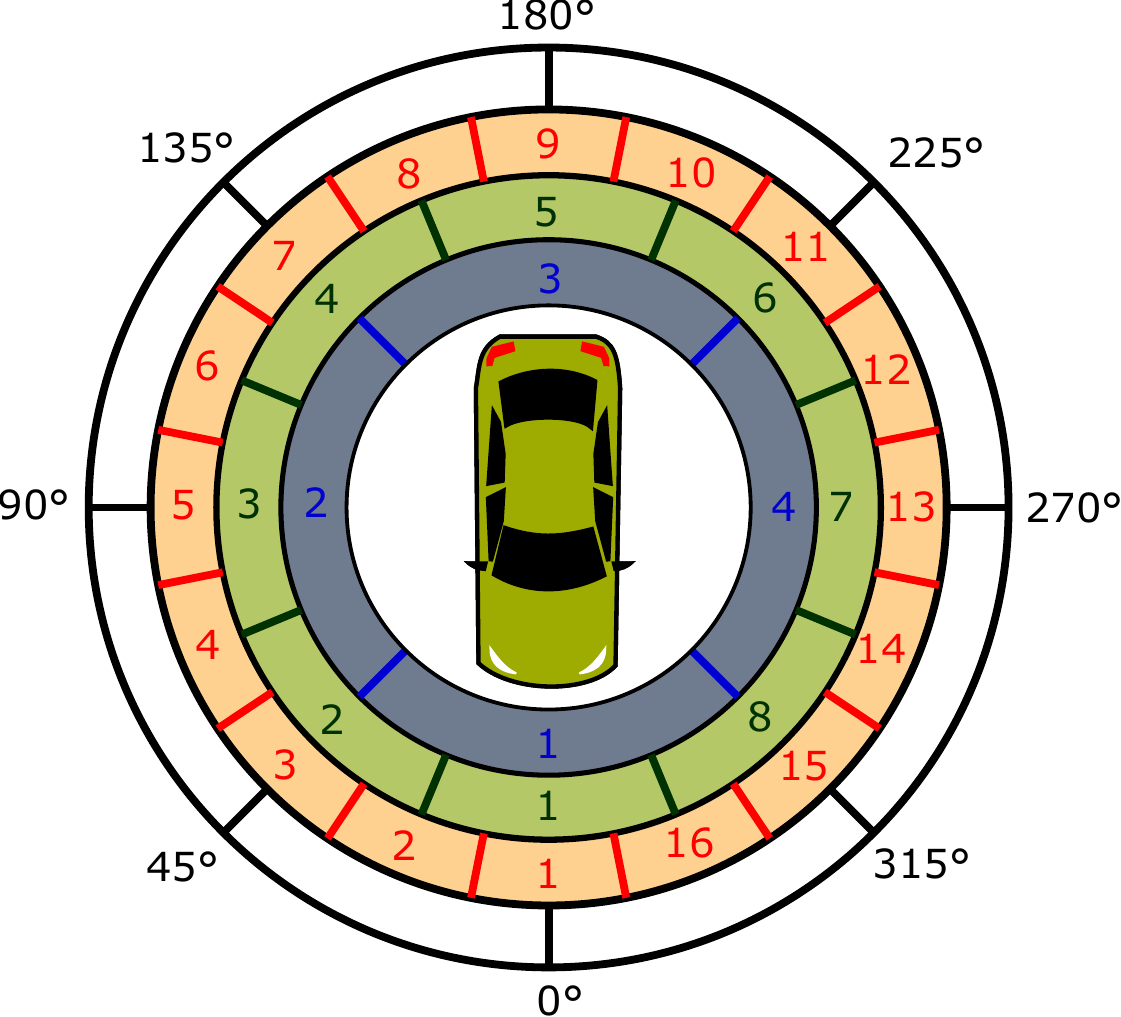}\label{fig:ViewpointClasses}} 

\caption{Definition of the viewpoint angle (a) and of the hierarchical viewpoint classes (b).
}
\label{fig:ViewpointBranch} 
\end{figure}

The viewpoint branch is added to the CNN in order to derive a probability distribution $\Pi_{\vartheta}$ for the vehicle's viewpoint $\vartheta$, which can be incorporated as prior information about the vehicle orientation into the probabilistic approach for model fitting. 
Since the direct prediction of the vehicle orientation $\theta$ from the detection window is not possible without the additional knowledge of the location of that window in the image, we train the \textit{viewpoint branch} to predict a a probability distribution $\Pi_{\vartheta}$ for the vehicle viewpoint $\vartheta$.
As depicted in Fig.~\ref{fig:ViewpointDef}, the viewpoint defines the aspect under which the vehicle is seen. 
Given the direction of the image ray $\rho$ to the 3D centre of the vehicle, the vehicle orientation $\theta$ can directly be computed from the viewpoint via 
\begin{equation}
\theta = 180^\circ-\vartheta - \rho. \label{eq:OriViewpoint}
\end{equation}
The \textit{viewpoint branch} performs a hierarchical classification of the viewpoint which is discretised in hierarchical viewpoint bins as depicted in Fig.~\ref{fig:ViewpointClasses}.
By merging the predictions of the individual classification heads using the probabilistic averaging layer, the final probability distribution $\Pi_{\vartheta}$ is obtained. 
For a detailed description of this branch, the reader is referred to \citep{Coenen2019}. 
%
%

\subsubsection{Keypoint/Wireframe branch} \label{sec:KeyptBranch}
This branch corresponds to a decoder network, upsampling the output of the input branch to the original input resolution using skip connections between corresponding layers of the encoder and decoder blocks.
Following the vehicle keypoint and wireframe definitions of Sec.~\ref{sec:ShapePrior}, the goal of this branch is to predict the presence of the appearance keypoints and the wireframe edges in the image.
Inspired by \citet{Newell2016} and \citet{KrishnaMurthy2017}, it is trained to produce one heatmap $\mathcal{H}^c_\mathcal{K}$ for every \textit{appearance} keypoint $c \in [1,C_A]$ in $\mathcal{K}_A$.
Additionally, the network is adapted to also output one heatmap $\mathcal{H}^w_\mathcal{W}$ for each of the wireframe definitions associated to one of the four sides $w \in \{\text{front}, \text{back}, \text{left}, \text{right}\}$.
A detailed definition of the \textit{appearance} keypoints and wireframe edges the network is supposed to predict has been given in Sec.~\ref{sec:ShapePrior}. 
The values at each pixel position of the resulting heatmaps correspond to a probability for the presence of the respective keypoint/wireframe edge at that position. 
The head of this branch consists of $C_A+4$ convolution layers producing the $C_A$ keypoint and the four wireframe heatmaps, using a sigmoid activation function to produce pixel-wise outputs in the interval $[0,1]$.

\subsubsection{Vehicle type branch} \label{sec:TypeBranch}
The \textit{vehicle type branch} proposed in this work is used to obtain a prediction of the vehicle type for the target vehicle shown in the input image window. 
This branch starts with a series of four convolutional layers followed by a max pooling layer.
Two fully connected layers are applied at the end of the branch and a softmax classification head delivers a class score for each of the vehicle type classes in $\mathcal{T}$ defined in Sec.~\ref{sec:ShapePrior}. 
It has to be noted that the categories of vehicle types do not have a clear definition and, therefore, the class boundaries are somewhat vague.
The car body configuration, which is determined by the layout of the engine, passenger and luggage volumes, as well as the number of pillars of a vehicle, i.e. the (almost) vertical supports of a car's roof and windows, are characteristics that can be used to distinguish different vehicle types.
However, ambiguities in the distinction of vehicle types exist and are therefore likely to be contained in the predictions of the vehicle type classification branch.
We therefore interpret the prediction scores as probabilities for the target vehicle to belong to the respective classes and gather them in a vector $\Pi_\mathcal{T}$, thus representing the probability distribution for the vehicle type.
The complete distribution is incorporated as prior information in the probabilistic model for vehicle reconstruction to introduce constraints on the deformations of the ASM used as shape prior, thus enabling the consideration of cases in which the type branch cannot clearly predict whether the vehicle belongs to one specific category or another. 

A detailed description of the training procedure of our CNN is given in Sec.~\ref{sec:ParamsAndTraining}

\subsection{Probabilistic model for vehicle reconstruction} \label{sec:ProbModel}
\begin{figure*}[ht]
\centering
		\includegraphics[width=0.9\textwidth]{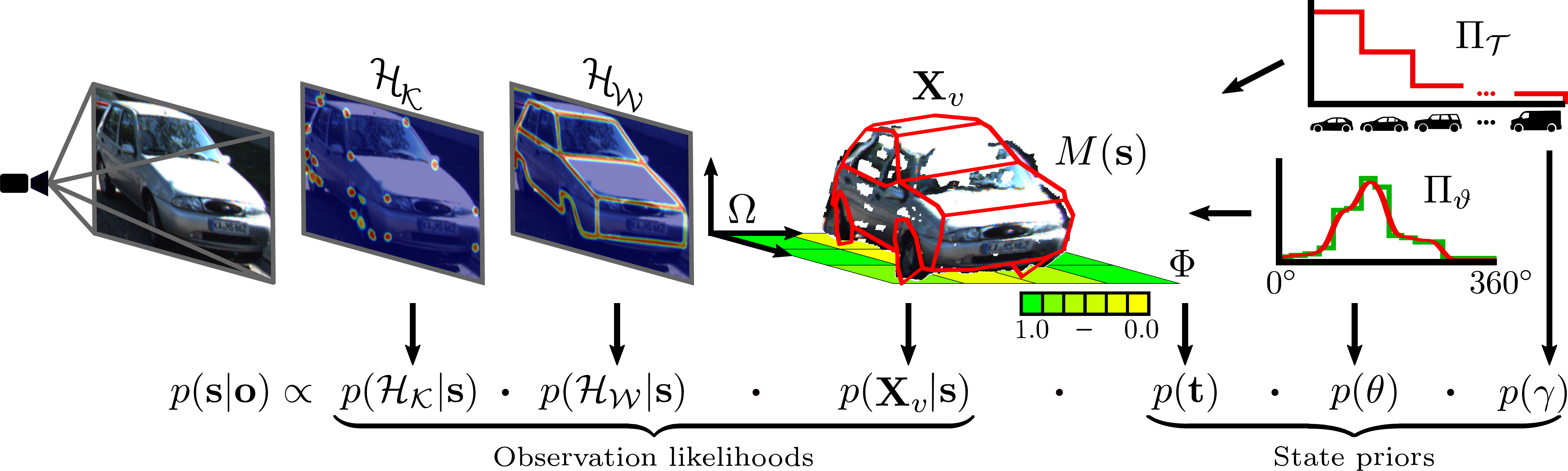}
	\caption{Visualisation of the probabilistic model. An ASM is  fitted to the reconstructed 3D points $\mathbf{X}_{v}$, the wireframe heatmaps $\mathcal{H}_\mathcal{W}$ and the keypoint heatmaps $\mathcal{H}_\mathcal{K}$. The probabilistic free-space grid map $\Phi$ and the probability distributions for the viewpoint and the vehicle type, $\Pi_{\vartheta}$ and $\Pi_\mathcal{T}$, are used to formulate regularisation terms for the state parameters $\mathbf{s}$.}
\label{fig:ProbabilisticFramework}
\end{figure*}

Given the vehicle detections $v \in \mathcal{V}$ and the observation vector $\mathbf{o} =(\mathbf{o}^{det}, \mathbf{o}^{cnn})$ associated to each detection, a vehicle model $M(\mathbf{s})$ is fitted to each detection by finding the optimal state variables $\hat{\mathbf{s}} = (\hat{\mathbf{t}}, \hat{\theta}, \hat{\gamma})$ for position, orientation and shape. 
For that purpose, a probabilistic model is formulated that simultaneously fits the surface of the 3D ASM to the 3D points, matches model keypoints to the detected keypoints and aligns the model wireframe to the wireframes inferred by the CNN. The inferred scene knowledge in form of the free-space grid map as well as the probability distributions for orientation and the vehicle type are incorporated in the probabilistic model as state priors.
Using a probabilistic formulation, the optimal state $\hat{\mathbf{s}}$ can be derived by maximising the posterior
\begin{equation}
 p(\mathbf{s}|\mathbf{o}) \propto p(\mathbf{o}|\mathbf{s}) \cdot p(\mathbf{s}) \rightarrow \max.
\end{equation}
In this work, the likelihood $p(\mathbf{o}|\mathbf{s})$ is factorised by individual likelihood terms, jointly incorporating both, 2D and 3D information derived from the stereo pairs and the CNN described in Sec.~\ref{sec:CNN}. 
The prior acts as a regularisation term on the state parameters and is factorised by one individual factor for each group of parameters, namely for position, orientation and shape, respectively. 
\begin{equation}
p(\mathbf{s}|\mathbf{o}) \propto \underbrace{p(\mathbf{X}_{v}|\mathbf{s}) \cdot p(\mathcal{H}_\mathcal{K}|\mathbf{s}) \cdot p(\mathcal{H}_\mathcal{W}|\mathbf{s})}_\text{Observation likelihood} \cdot \underbrace{p(\mathbf{t}) \cdot p(\theta) \cdot p(\gamma)}_\text{State prior}. \label{eq:posterior}
\end{equation}
This model is visualised in Fig.~\ref{fig:ProbabilisticFramework}.
The likelihood is composed of a \textit{3D likelhihood} $p(\mathbf{X}_{v}|\mathbf{s})$ based on the 3D vehicle points $\mathbf{X}_{v}$ as well as the \textit{keypoint} and \textit{wireframe likelihoods} $p(\mathcal{H}_\mathcal{K}|\mathbf{s})$ and $p(\mathcal{H}_\mathcal{W}|\mathbf{s})$, which are based on the keypoint and wireframe heatmaps $\mathcal{H}_\mathcal{K}$ and $\mathcal{H}_\mathcal{W}$, respectively.
The state priors for \textit{position}, \textit{orientation} and \textit{shape} are derived based on the probabilistic free-space grid map $\Phi$, the probability distribution for the viewpoint $\Pi_{\vartheta}$ and the prediction for the vehicle type $\Pi_\mathcal{T}$, respectively.
An energy function $E$ is defined which corresponds to the negative logarithm of the posterior of Eq.~\ref{eq:posterior} and which is minimised to find the optimal state parameters.
The logarithmic formulation of the individual likelihood and prior terms is the same as in \citep{Coenen2019}, except that we added the \textit{shape prior} term $p(\gamma)$. To make the paper self-contained, we explain all terms in the following sections.
%
%

\subsubsection{3D likelihood} \label{sec:3DLikelihood}
This likelihood term is based on the distance of the 3D points $\mathbf{X}_{v}$ reconstructed from image points representing the surface of a vehicle (cf.\ Sec.~\ref{sec:VehicleDetection}) from the Model surface $M_\Delta$ of the model $M(\mathbf{s})$:
\begin{equation}
\log p(\mathbf{X}_{v}|\mathbf{s}) = -\frac{1}{P} \sum_{x\in \mathbf{X}_{v}} \frac{\mathcal{L}_\text{H}(x, M_\Delta)}{2\sigma_x^2}.
\label{eq:E3D}
\end{equation}
Here, $P$ is the overall number of 3D points in $\mathbf{X}_{v}$ and $\sigma_x$ is the depth uncertainty of the individual 3D point $x$, which is determined by applying error propagation to stereo triangulation and using an uncertainty for the disparity estimate $\sigma_{\text{disp}}$ of 1~[px].
To add robustness against possible outliers remaining in $\mathbf{X}_{v}$ we use the Huber loss \citep{Huber1964} to calculate the distance $\mathcal{L}_\text{H}(x, M_\Delta)$:
\begin{equation}
\mathcal{L}_\text{H}(x, M_\Delta) = 
\begin{cases}
d(x,M_\Delta)^2 & \text{if}\ d(x,M_\Delta) \leq \sigma_x, \\
2\sigma_x \cdot d(x,M_\Delta) - \sigma_x^2 & \text{otherwise}.
\end{cases}
\end{equation}
The Huber loss is more robust against outliers compared to the quadratic distance $d(x,M_\Delta)^2$ of a 3D point $x$ from the model surface $M_\Delta$. 
To determine $d(x,M_\Delta)$, the distance of the 3D point $x$ to every triangle in $M_\Delta$ is calculated and the distance to the model surface is defined as the smallest distance found. 
Minimizing this term fits the 3D ASM to the 3D point cloud.

\subsubsection{Keypoint likelihood} \label{sec:keypointLikelihood}
The \textit{keypoint likelihood} is based on the idea that, when backprojected to the image planes, the keypoints of the ASM representing the true vehicle should be supported by keypoint detections inferred from the image data (i.e.\ by high probabilities in the keypoint heatmaps) at or close to the backprojected keypoint positions.
For this purpose, the keypoints $\mathcal{K}_A$ of the model $M(\mathbf{s})$ are backprojected to the stereo images, resulting in a list of $c=[1,C_A]$ image points $\mathbf{u}^{l/r}_c$ for both, the left ($l$) and right ($r$) stereo images, respectively. 
Note that the CNN described in Sec.~\ref{sec:CNN} is trained using only the keypoints being visible in the image and not being (self-)occluded.
Consequently, the network is able to detect  visible keypoints only, which is the reason why self-occlusion caused by the vehicle model itself is considered in the calculation of the keypoint likelihood.
Ray tracing techniques can be used to reason about visible and self-occluded model keypoints.
However, these techniques usually are computational expensive.
Instead, the rigid topology of the ASM surface and the street-level acquisition setup allow the construction of a look-up table, storing the set of visible model keypoints for every viewpoint $\vartheta \in [0,360^\circ]$.
Using this look-up table, a boolean variable $\delta_c$ is associated to every image keypoint, with $\delta_c =  1$ if the keypoint $c$ is visible and inside of the detection bounding box and $\delta_c = 0$ otherwise.
The total number of visible keypoints is $U=\sum_{c=1}^C \delta_c$.
The keypoint likelihood is calculated by
\begin{equation}
\log p(\mathcal{H}_\mathcal{K}|\mathbf{s}) = -\frac{1}{2U} \sum_{i \in \{l,r\}} \sum_{c=1}^{C_A} \delta_c \cdot \log \left(1- {^i}\mathcal{H}_\mathcal{K}^c(\mathbf{u}_c^i)\right).
\end{equation}
Here, ${^i}\mathcal{H}_\mathcal{K}^c(\mathbf{u}_c)$ denotes the output of the heatmap for the keypoint $c$ at the location ${^i}\mathbf{u}_c$ in image $i\in \{l,r\}$.
Minimizing this term fits the 3D ASM to the keypoints predicted in both images.

\subsubsection{Wireframe likelihood} \label{sec:wireframeLikelihood}
The \textit{wireframe likelihood} is based on a measure of similarity between the backprojected edges of the model wireframe $M_\mathcal{W}$ and the wireframe heatmaps $\mathcal{H}_\mathcal{W}$ inferred from the CNN.
To this end, we backproject the visible parts of the wireframe subsets $w \in \{\text{front}, \text{back}, \text{left}, \text{right}\}$ to the left and right images, resulting in binary wireframe images ${^l}I_\mathcal{W}^w$ and ${^r}I_\mathcal{W}^w$ with entries of 1 at pixels that are crossed by a wireframe edge in subset $\mathcal{W}$ and 0 everywhere else. 
To consider differences between the real image wireframe positions and the model wireframe caused by generalisation effects of the vehicle model, the wireframe images are blurred using a Gaussian filter.
The size of the filter is defined according to the backprojection uncertainty of the model keypoints given the generalisation error of the ASM quantified by an uncertainty $\sigma_M$ (set to 10\,cm in this work). 
The applied Gaussian filter is defined according to the resulting backprojection uncertainties, leading to a stronger blurring effect when the vehicle is close to the camera and vice versa.

Given the heatmaps $\mathcal{H}_\mathcal{W}$ and the blurred images of the backprojected model wireframe $I_\mathcal{W}$, the wireframe likelihood is calculated according to
\begin{equation}
\log p(\mathcal{H}_\mathcal{W}|\mathbf{s}) = -\frac{1}{2} \sum_{i \in \{l,r\}} \sum_w \log\left(1-BC({^i}I_\mathcal{W}^w, {^i}\mathcal{H}_\mathcal{W}^w)\right) 
\end{equation}
where we use the Bhattacharyya coefficient $BC(\cdot, \cdot)$ \citep{Bhattacharyya} as a similarity measure between the blurred wireframe images and the wireframe heatmaps.
This term will become large if the backprojected wireframes correspond well to the wireframes predicted by the CNN.

\subsubsection{Position prior}
The \textit{position prior} is derived from the probabilistic free-space grid map $\Phi$. It is calculated based on the amount of overlap between the minimum enclosing 2D bounding box $M_\mathcal{B}$ of the model $M(\mathbf{s})$ on the ground plane and the free-space grid map cells $\Phi_g$ with $g = [1,G]$ given their probability $\rho_g$ of being free space (\textit{unknown} cells are disregarded by setting $\rho_g = 0$):
\begin{equation}
\log p(\mathbf{t}) = \frac{\lambda_\Phi}{A_\mathcal{B}} \sum_{g=1}^G \log(1-\rho_g) \cdot o(M_\mathcal{B}, \Phi_g).
\label{eq:Efree}
\end{equation}
$A_\mathcal{B}$ is the area of the model bounding box. The function $o(\cdot,\cdot)$ calculates the overlap between the model bounding box and a cell $\Phi_g$ using the surveyor's area formula \citep{Shoelace}. 
If there is no intersection between $M_\mathcal{B}$ and $\Phi_g$, $o(\cdot , \cdot)$ returns zero.
As the free-space grid map is derived from the reconstructed 3D points, a factor $\lambda_\Phi$ is introduced to transfer the uncertainties of the 3D points to the calculation of the position prior.
To this end, the factor $\lambda_\Phi = \min(1, \frac{l_{\Phi}}{\sigma_x})$ is used as a weight of this term based on the grid cell size $l_\Phi$ and the depth uncertainty $\sigma_x$ of a reconstructed point $x$ in the distance of the model $M(\mathbf{s})$. This prior penalises models that are partly or fully located in areas which are observed as not being occupied by 3D objects. 
Particularly, this prior establishes the constraint that free-space between the camera and the point cloud cannot be occupied by the model. 

\subsubsection{Orientation prior}
To calculate the \textit{orientation prior} for the model $M(\mathbf{s})$, the probability distribution $\Pi_{\vartheta}$ for the vehicle viewpoint inferred by the multi-branch CNN is used.
The viewpoint $\vartheta_M(\theta)$ is computed for the model $M(\mathbf{s})$ from the model orientation $\theta$ using the relation in Eq.~\ref{eq:OriViewpoint}.
The image ray direction $\rho$ is derived from the ray connecting the camera projection center and the centre of the vehicle model $M(\mathbf{s})$. The \textit{orientation prior} is calculated according to
\begin{equation}
\log p(\theta) = \log \Pi_{\vartheta}(\vartheta_M(\theta)) + \log \left(\frac{1+\cos(\vartheta_{\text{CNN}} - \vartheta_M(\theta))}{2}\right).
\end{equation}
$\Pi_\vartheta(\vartheta_M(\theta))$ denotes the probability for the angle $\vartheta_M(\theta)$ according to the output of the viewpoint classification branch of the CNN.
As incorrect viewpoint classifications can be assumed to appear especially between neighbouring viewpoints, this term alone is prone to cause small orientation errors. This is why additionally, the cosine distance of the most likely viewpoint $\vartheta_{CNN}$ predicted by the vehicle CNN and the model viewpoint $\vartheta_M$ is considered in this prior.

\subsubsection{Shape prior} \label{sec:ShapePriorTerm}
The new \textit{shape prior} formulation proposed in this paper is based on the probability distribution $\Pi_\mathcal{T}$ for the vehicle types predicted by the CNN and acts as regularisation term for the shape of the model $M(\mathbf{s})$.
Using
\begin{equation}
\log p(\gamma) = -\frac{1}{n_s}\sum_{\tau \in \mathcal{T}} \sum_{s=1}^{n_s} \Pi_\mathcal{T}^\tau \cdot \frac{(\gamma_s^\tau - \gamma_s)^2}{2\sigma_s^2}, \label{eq:ShapePrior}
\end{equation}
the \textit{shape prior} term penalises deviations of the $n_s$ considered model shape parameters $\gamma_s$ from the predicted modes $\gamma^\mathcal{T}_s$ of the individual vehicle types.
In this prior formulation, the deviations from each mode are weighted according to the probability scores $\Pi_\mathcal{T}$ predicted by the CNN with $\sum \Pi_\mathcal{T}^\tau = 1$.
The parameter $\sigma_s$ represents the ASM standard deviation of deformation, i.e.\ the square root of the eigenvalue, in the direction of the eigenvector associated to the $s^\text{th}$ shape parameter as described in Sec.~\ref{sec:ShapePrior}.

As it is reasonable to assume the ASM shape parameters of vehicles not to follow a uni-modal distribution but rather a multi-modal distribution with each mode representing one vehicle category, penalising deviations from the overall mean shape, as is usually done in the literature \citep{Zia2,Engelmann}, is an unfavourable procedure. The category aware \textit{shape prior} formulation proposed here gives a more realistic and detailed constraint on the shape parameters. The confidence awareness in the prior term, achieved by considering the probability scores $\Pi_\mathcal{T}$ for all distinguished categories, reduces the sensitivity of the \textit{shape prior} to potential uncertainties in the vehicle type predictions of the CNN, e.g. caused by the vague definition of the car type classes discussed in Sec.~\ref{sec:TypeBranch}.

\subsubsection{Inference}
To find the optimal pose and shape parameters for each detected vehicle, the energy function derived from Eq.~\ref{eq:posterior} is minimised.
As this function is non-convex and discontinuous due to the changing visibility of keypoints/wireframes caused by self-occlusion, a sequential Monte Carlo sampling procedure is applied to approximately determine the parameter set for which the energy function becomes minimal.
To this end, the target parameters are sampled to generate model particles for the vehicle ASM. 
Starting from one or more initial parameter sets, a number of particles $n_p$ are generated in each iteration $j = 1,...n_{it}$ by jointly sampling the pose and shape parameters from a uniform distribution centred at the preceding parameter values. 
For the resampling step, the energy for every particle is calculated in each iteration $j$ and the $n_b$ highest weighted particles are introduced as initial seed particles for the next iteration. 
In each iteration, the size of the interval from which the parameters are sampled is reduced. 
In the following paragraphs, more details are given.\\

\textbf{Initialisation:} A common way for initialisation is to define the initial particles $\mathbf{s}_0^i$ with $i=1,...,n_p$ using the prior distribution $p(\mathbf{s})$.
With $p(\mathbf{s}) = p(\mathbf{t}) \cdot p(\theta) \cdot p(\gamma)$ the prior distribution heavily depends on the predictions of the CNN for the orientation and the shape.
The most likely particle $\mathbf{s}_0^1 = (\mathbf{t}_0^1, \theta_0^1, \gamma_0^1)$ is determined from $p(\mathbf{s})$ by setting $\theta_0^1$ to the most likely orientation given $\Pi_{\vartheta}$ and $\gamma_0^1$ to the most likely shape $\gamma^\tau$ given $\Pi_\mathcal{T}$.
The initial translation vector $\mathbf{t}_0^1$ is defined as the bounding box centre of the minimum 2D bounding box enclosing the 2D projections of the 3D vehicle points $\mathbf{X}_{v}$ to the ground plane.
The particles $\mathbf{s}_0^i$ are sampled using an uniform distribution for position, orientation and shape, respectively, centred at the most likely particle $\mathbf{s}_0^1$. The initial interval boundaries are set to $\pm 180\grad$ and $\pm 1.5$~[m] for the orientation and position parameters, respectively, and to $\pm 3$ for the shape parameters. By choosing $\pm 180\grad$ as interval for the orientation angle, particles are allowed to take the whole range of possible orientations in the first iteration to be able to deal with incorrect initialisations, thus gaining robustness against initialisation errors. \\

\textbf{Resampling:} To resample the particles in each iteration $j = 1,...,n_{it}$ the energy $E$ is calculated for each preceding particle.
The best $n_b$ particles, i.e.\ particles with the lowest energy, are introduced as seed particles for the resampling step.
For each seed particle, an equal number of $N_j = \frac{n_p}{n_b}$ of offspring particles $\mathbf{s}_j$ are drawn from a uniform distribution centred at the respective seed particle. To encourage convergence of the applied Monte Carlo sampling, the range for the respective parameters used to draw the particles from the uniform distributions is reduced by a factor $0.85^j$ in iteration $j$.
As a consequence, the initial parameter range is reduced by the factor $0.85^{n_{it}}$ in the last iteration $j=n_{it}$. 
By forwarding multiple particles, the preservation of particles potentially belonging to different minima is enabled. As a consequence, the risk of getting stuck in a local minimum is reduced, which allows to deal with multi-modal distributions and local minima in the objective function.\\

\textbf{Final result:} The final values for the target parameters of pose and shape are defined after the last iteration and are set to the parameters of the particle achieving the lowest energy within the particle set of the final iteration. 
\section{Test data and test setup} 

\subsection{Test data} \label{sec:TestData}
The empirical evaluation of the proposed method is conducted on two data sets. 
Both data sets consist of stereo image pairs which were acquired by a synchronised and calibrated stereo camera rig, placed on a mobile platform, while driving in regular traffic on public roads in urban environments. 
One data set is taken from the publicly available KITTI benchmark suite \citep{KITTI} and will therefore be referred to as KITTI data throughout this paper.
The second data set was presented in \citep{Coenen2019} and has been made publicly available\footnote{https://doi.org/10.25835/0078519} \citep{Coenen2020}. 
%
\begin{figure}[H] 
\centering
\subfloat[KITTI] {\includegraphics[width=0.99\columnwidth]{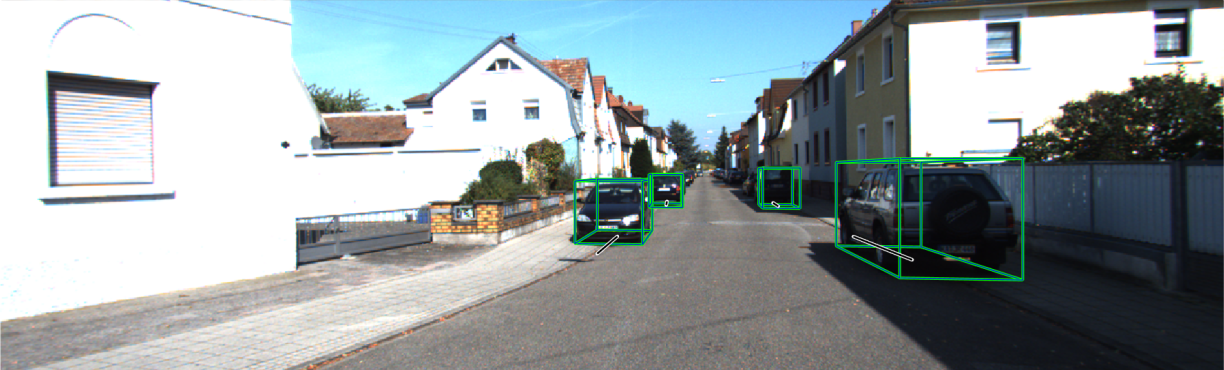}\label{fig:KITTIExample}} \linebreak
\subfloat[ICSENS] {\includegraphics[width=0.99\columnwidth]{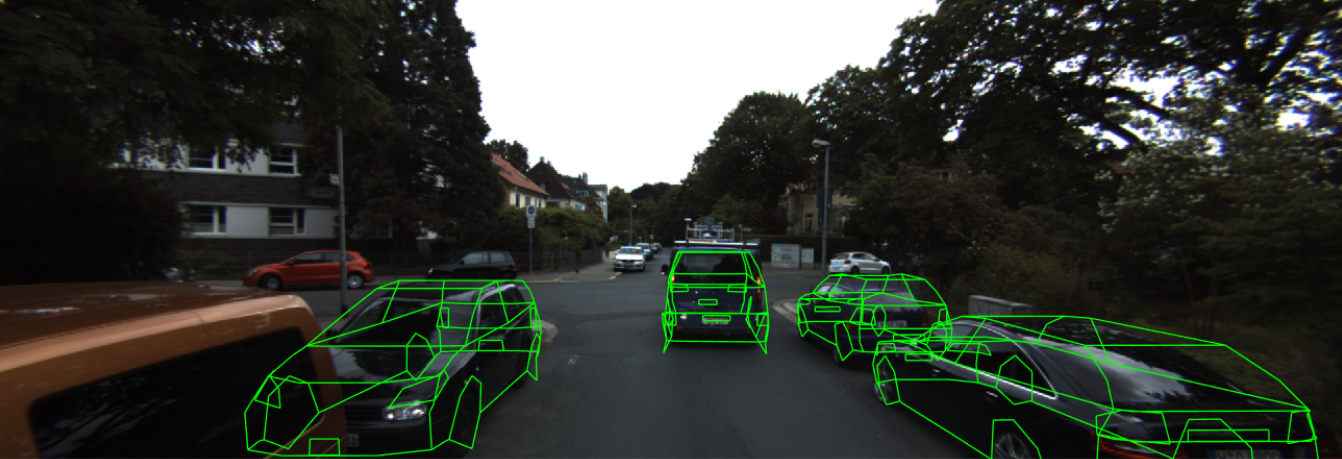}\label{fig:MapathonExample}} 
\caption{Examples from the test data sets. In (a), an example from the KITTI benchmark is shown, providing 3D reference bounding boxes. In (b), an example of the ICSENS data set is shown, which provides fitted 3D CAD models as reference.}
\label{fig:DatasetExamples} 
\end{figure}
%
It will be referred to as ICSENS (\textit{Integrity and Collaboration in dynamic SEnsor NetworkS}, \citep{icsens2018}) data set in the remainder of this paper.
Regarding the KITTI data, the 3D object detection benchmark is used for the evaluation in this paper.
It consists of 7481 stereo image pairs and provides the 3D object location and the orientation for every vehicle in the form of manually fitted oriented 3D bounding boxes. It distinguishes three levels of difficulty (easy, moderate and hard), which mainly depend on the level of object occlusion and truncation.
In this paper, 260 of the training set images are used for training and the remaining images are used for evaluating the proposed approach.
The KITTI test set consists of 7518 stereo images for which no reference is provided. 
An official evaluation on the test set using the official KITTI metrics is used to compare the performance of the proposed method to the results of related methods which are reported in the KITTI leaderboard\footnote{http://www.cvlibs.net/datasets/kitti/eval\_3dobject.php}.
The ICSENS data set consists of a total of 1000 stereo image pairs recorded in the context of \citep{icsens2018}.
In contrast to the KITTI dataset, which only delivers oriented 3D bounding boxes
as references, the ICSENS dataset delivers the reference shape and the reference vehicle type in addition to its 3D pose. 
To this end, we manually fitted the most similar model out of a set of vehicle CAD models to the individual vehicles of the ICENS data set, using the 3D point cloud obtained from stereo matching and the back-projected wireframes to assess whether a reference model was correct.  
We also distinguish between \textit{easy} (fully visible)  and \textit{difficult} (occluded or truncated) vehicles. 
A visual comparison of the provided references for the KITTI and the ICSENS data can be seen in Fig.~\ref{fig:DatasetExamples}. 
Note that the quality of the reference will be affected by the depth errors of the reconstructed 3D points in the same way as the reconstructed models.

\subsection{Parameter setting and training} \label{sec:ParamsAndTraining}
We select the side length $l_\Phi$ of the free-space grid cells to be 25 cm. 
For inference, we define the number of particles to be $n_p$ = 200; the number of iterations $n_\text{it}$ and the number of offspring particles $n_b$ are both set to 10. 
The category-aware Active Shape Model, which is proposed as a shape prior in this work (cf.\ Sec.~\ref{sec:ShapePrior}) requires the definition of a set of type classes $\mathcal{T}$ and the availability of 3D keypoint annotations for a set of training exemplars.
Using $\mathcal{T} = $~\textit{\{Compact\ Car,\ Sedan,\ SUV,\ Estate\ Car,\ Sports\ Car,\ Truck,\ Van\}}, seven vehicle categories are distinguished  (cf.\ Fig.~\ref{fig:ASM}). 
To learn the ASM, $C_\mathcal{K} =144$ 3D keypoints were manually labelled on a set of 36 different CAD vehicle models collected via \textit{Google's 3D Warhouse}\footnote{https://3dwarehouse.sketchup.com} and belonging to one of the considered vehicle types. Each model differs from the other models in shape and, consequently, also in its 3D extents. 
Tab.~\ref{tab:StatisticsASM} shows the statistics that are obtained from the variations in length, width, and height of the CAD models used in the training set in this work.
It shows the mean value, standard deviation (std. dev.) as well as the minimum and maximum values for length, width, and height of the vehicles.
Obviously, with a standard deviation of 0.40\,m, the largest variations are present w.r.t.\ the length of the vehicles.
With standard deviations of 0.20\,m and 0.10\,m, respectively, the variations in height and width are considerably smaller. 
A more detailed overview on the standard deviations of length, width, and height of the CAD vehicles for each of the considered vehicle types can be seen in Tab.~\ref{tab:StatisticsASMType}. 
It shows that the largest variations in all directions result for the class \textit{Van}. The class \textit{Sedan} exhibits the smallest intra-class variations.
\begin{table}[H]
	\centering
	\caption{Statistical properties for the vehicle extents of the CAD training set used to learn the ASM.}
		\begin{tabular}{|l|c c c c|}\hline
				[m]		& mean & std. dev. & max & min \\ \hline
		length 	& 4.35	& 0.40 	& 5.70 & 3.55 \\
		width 	& 1.80 	& 0.10 	& 2.34 & 1.65 \\
		height	& 1.49 	& 0.20 	& 2.12 & 1.11 \\ \hline
						
		\end{tabular}	
\label{tab:StatisticsASM}
\end{table}  
\begin{table*}[ht]
	\centering
	\caption{Standard deviations of the vehicle dimensions differentiated by the individual vehicle types.}
		\begin{tabular}{|l|c c c c c c c|}\hline
					[m]	& \textit{Compact Car} & \textit{Estate} & \textit{Sedan} & \textit{SUV	} & \textit{Van} & \textit{ Sports Car} & \textit{Truck}\\ \hline
		length 	& 0.21	& 0.24	& 0.19	& 0.39	& 0.82	& 0.22	& 0.26 \\
		width 	& 0.05	& 0.07	& 0.04	& 0.07	& 0.28	& 0.04	& 0.19 \\
		height	& 0.10	& 0.06	& 0.05	& 0.14	& 0.28	& 0.13	& 0.12 \\ \hline						
		\end{tabular}	
\label{tab:StatisticsASMType}
\end{table*}  

\begin{figure}[H]
\centering
\includegraphics[width=0.95\columnwidth]{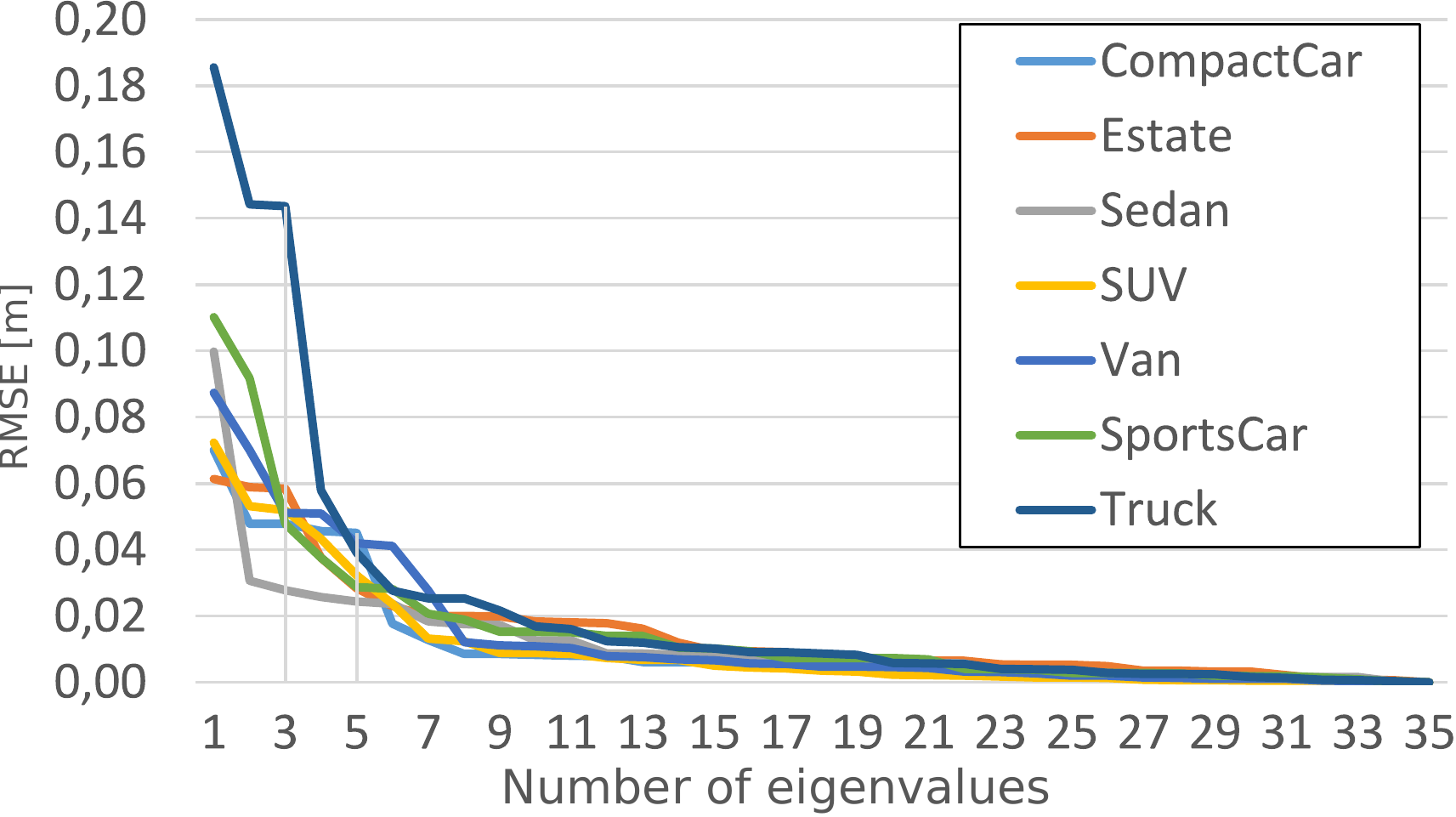} 
\caption{RMSE of the ASM fitting to the individual considered vehicle types.}
\label{fig:ASMRMSE} 
\end{figure}

The set of \textit{appearance keypoints} $\mathcal{K}_A \in \mathcal{K}$ considered in this work contains $C_A=36$ individual keypoints and corresponds to the keypoints used in \citep{Zia2}.
As explained in Sec.~\ref{sec:ShapePrior}, a shape basis is learned for each of the considered vehicle types.
In Fig.~\ref{fig:ASMRMSE}, the fitting error of the overall ASM to the mean shape $\mathbf{m}^\tau$ of the vehicle types $\tau$ resulting from Eq.~\ref{eq:ASMModeLearning}, i.e. the root mean square error (RMSE) of the keypoint coordinates, is shown as a function of the number of principal components used.
The RMSE represents the generalisation error of the ASM caused by using a restricted number of eigenvaectors. For the number of eigenvalues and eigenvectors to be considered in the ASM during the experiments we choose $n_s = 3$, which we found to be a proper tradeoff between the complexity of the model and the quality of the model fit.
While the RMSE of the type $Truck$ is 0.14\,m when using the first three principal components, the RMSE of the remaining vehicle types is smaller than 0.06\,m in all cases.
%
%

To train the proposed multi-branch CNN, images of vehicles cropped by the tightly enclosing bounding box of the vehicles are used.
We applied the following training strategy in our experiments.
The input branch (cf.\ Fig.~\ref{fig:VehicleCNN}) is initialised from its corresponding layers of the VGG19 network \citep{VGG19}, pre-trained on ImageNet \citep{ImageNet}, and is frozen during the training procedure.
The remaining convolutional layers are initialised using the \textit{He} initialiser \citep{He2015}.
For the presented experiments, the \textit{type branch} is trained separately from the \textit{viewpoint} and the \textit{keypoint/wireframe} branches.
To train the individual output branches, two different data sets are used. 
For the joint training of the \textit{viewpoint} and the \textit{keypoint/wireframe} branches, training images of vehicles are required, including reference information of the vehicle's viewpoint as well as the image coordinates of the \textit{appearance keypoints}. 
To this end, the 260 KITTI images mentioned in Sec.~\ref{sec:TestData} are used.
The 2D image reference bounding boxes as well as the reference viewpoint angles are provided as annotations of the data set \citep{KITTI}. 
The authors of \citep{Zia3} labelled the 36 different \textit{Appearance} keypoints in this subset of images and made the annotations publicly available.
Together with the reference viewpoint angles, their keypoint annotations are used to train the \textit{keypoint/wireframe branch} and the \textit{viewpoint branch} for the experiments conducted in this paper, while the \textit{vehicle type branch} is frozen. 
For more details on the training procedure of the \textit{viewpoint} and \textit{keypoint/wireframe branches} we refer the reader to \citep{Coenen2019}. 
Note that the KITTI images used for training are not used for the evaluation.
After training both branches, the \textit{vehicle type branch} is trained independently from the remaining branches using the data set provided by \citep{Yang2015}, which contains images of vehicles including bounding box and vehicle type annotations. 
In this context, the reference vehicle type (one out of $\mathcal{T}$) is assumed to be available for each training sample.
In \citep{Yang2015}, twelve different vehicle type classes are distinguished. To map the provided classes to the type definitions used in this work, some classes are merged as shown in Tab.~\ref{tab:TypeClassMerge}.
The categorical cross-entropy is used as loss function to train the \textit{vehicle type branch}.
The entire network is trained using the Adam optimizer \citep{Adam2015}, a variant of stochastic mini-batch gradient descent with momentum, using the exponential decay rate for the $1^\text{st}$ moment estimates $\beta_1 = 0.9$ and for the $2^\text{nd}$ moment estimates $\beta_2 = 0.999$.  
A mini-batch size of $N=50$ and an initial learning rate of $\nabla = 10^{-4}$ are applied.
To improve training, the learning rate is decreased by a factor of $10^{-1}$ after 5 epochs with no improvement in the validation loss.  
Furthermore, batch normalisation \citep{BatchNormal2015} is used and Dropout \citep{Dropout} is applied to the fully-connected layers with a rate of 0.5.
Data augmentation is applied to the training data by horizontally flipping the training images, consequently adapting the viewpoint classes and keypoint/wireframe labels accordingly. The classes for the vehicle type remain unchanged. 
During training, regions corresponding to 0 and 20\% of the bounding box width are randomly clipped away from the image to  simulate occlusions. We expect this to help the network to derive suitable features even if the vehicles are not fully visible.
Further, random gamma corrections with gamma in the range of $[0.1, 2.0]$ are applied to the training images to enforce robustness against radiometric differences, e.g. due to illumination conditions, background-foreground contrast, shadowing, etc. 
\begin{table*}[ht]
	\centering
	\caption{Vehicle type class definition made in this paper and the associated classes defined in \citep{Yang2015}.}
		\begin{tabular}{l|l l l l l l l}
		Ours 							& \textit{Compact Car} & \textit{Sedan} & \textit{SUV} & \textit{Estate Car} & \textit{Sports Car} & \textit{Truck} & \textit{Van} 		\\ \hline
		Yang et al.			 	& Hatchback		& Sedan 		& SUV 				& Estate car 	& Sports car 		& Pickup 	& MPV 		\\ 
		(2015)						& 				 		& Fastback	&	Crossover		& 						& Convertible 	& 				& Minibus \\
											&							&						&							&							& Hardtop conv. & 				& 				\\ 
		
		\end{tabular}	
\label{tab:TypeClassMerge}
\end{table*}  

\subsection{Test setup}
The \textit{vehicle type branch} introduced in this work is used to derive a prior distribution for the vehicle type to be incorporated in the proposed probabilistic model to regularise the vehicle shape. 
An evaluation of the performance of this classification branch allows to draw conclusions about the suitability of the proposed CNN for the derivation of shape prior information. 
The overall accuracy (OA) is used as evaluation criterion, which is computed according to the ratio of the number of correctly classified vehicles and the total number of vehicles, and therefore represents the overall proportion of correct classifications in [\%].
As explained in Sec.~\ref{sec:TypeBranch}, the class boundaries between some the considered vehicle type classes are not clearly defined.
The vague definition of the class boundaries is expected to be reflected by a broad distribution of the confidence scores predicted by the vehicle type branch over the concerned classes.
To account for this effect in the evaluation, different values of OA are reported: The Top-1 OA, which reports the percentage of correct classifications that are obtained by using the class exhibiting the highest confidence score as the predicted one.
In addition, the Top-2 and Top-3 OA values are also analysed.
To obtain the Top-2 and Top-3 accuracies, a sample is considered to be a true positive if the reference class is among the two or three classes having the highest confidence scores, respectively.
For the evaluation, the ICSENS data set is used, because it contains a reference for the vehicle type.

To asses the impact of the components of the probabilistic model proposed in this paper, different variants for the probabilistic model are defined, each of them considering a different set of likelihood and prior terms as described in the following paragraphs. 

\textbf{Base:} 
In this variant, the baseline setting is defined in which only the \textit{3D likelihood} term is considered for model fitting.
The \textit{3D likelihood} $p(\mathbf{X}_{v}|\mathbf{s})$ is chosen to define the baseline model because it is exclusively based on the ASM and the reconstructed 3D points and, therefore, this variant does not require any supervised learning.
The observations generated by the CNN (keypoint and wireframe heatmaps) are not considered here.
Note that in this variant, the regularisation of shape deformations by the category-aware shape prior is omitted. 
However, constraints on the shape prior still have to be introduced to avoid unconstrained shape variations of the ASM, which may result in geometrically invalid vehicle shapes.
In this setting the shape parameters are chosen to be regularised by penalising deviations from the mean ASM shape, as it is also done in comparable related methods, e.g.\ \citep{Zia2, Engelmann}.
By doing so, the category-aware shape prior formulation proposed in this paper (Eq.~\ref{eq:ShapePrior}) is replaced by
\begin{equation}
\log p(\gamma) = -\frac{1}{n_s} \sum_{s=1}^{n_s} \left(\frac{\gamma_s}{2\sigma_s}\right)^2. \label{eq:ShapePriorUninformed}
\end{equation}

\textbf{Base+S:} 
In this setting, the \textit{shape prior} term $p(\gamma)$ according to Eq.~\ref{eq:ShapePrior} is added to the model alignment based on the \textit{3D likelihood}. The potential benefit of the category-aware shape prior in comparison to the regularisation setting defined here can be analysed in this variant.

\textbf{Base+S+P+O:}
To  assess the potential of the complete state priors, the \textit{shape, position}, and \textit{orientation priors} are jointly considered as regularisers on the state parameters during alignment in this variant. 

\textbf{Base+K+W}
This setting considers all likelihood terms presented in this paper to assess their potential. 
To constrain the vehicle shape, Eq.~\ref{eq:ShapePriorUninformed} is applied to prevent the vehicle models from degenerating.

\textbf{Full:} 
The evaluation of the vehicle reconstructions using the complete probabilistic formulation of Eq.~\ref{eq:posterior}, referred to as the \textit{Full} model, assesses the quality of the results that can be obtained by the presented approach. 
The comparison of these variants is performed on the KITTI data set, while the \textit{Full} model is also evaluated on the ICSENS data. 

\textbf{Evaluation criteria:}
To evaluate the vehicle reconstruction, the resulting pose and shape of each fitted 3D vehicle model are compared to the reference. 

To evaluate the shape of the vehicle reconstructions, the vehicle dimensions is one criteria that can be used in case of the KITTI data, by using the reference 3D bounding boxes. 
To this end, average absolute errors are computed from the differences of the reference and the inferred length, width, and height of the vehicles for the KITTI dataset. 
In addition, the point clouds acquired by the Velodyne laserscanner, which were used to generate the ground-truth data of the KITTI data set, can be used to obtain a more detailed evaluation of the shape of the reconstructed vehicle model. 
To this end, we transform the laserscanner point cloud to the model coordinate system and extract the set of 3D laserscanner points lying inside the 3D reference bounding box of each vehicle. 
To assess the quality of the fitted models, for each vehicle we compute the RMSE of the distances of the laser points associated with a vehicle from the surface of the fitted ASM. 
The average and the median values computed from the RMSE of all vehicles in the data set are reported in the evaluation. 
In addition to the bounding box dimension metrics, the errors based on the Velodyne points give a more precise view of the deviations of the reconstructed  models from the true shape of the vehicles. 
Furthermore, whereas the bounding box metrics assess the quality of the full extent of the vehicle, i.e.\ including the self-occluded and therefore non visible parts, the error metrics based on the Velodyne points only assess the reconstruction quality of the vehicle parts which are visible to the camera and therefore do not consider the self-occluded and therefore ambiguous vehicle parts.

Instead of 3D bounding boxes, the ICSENS data provides CAD models as reference.
In order to compute the same evaluation metrics for the ICSENS data, the minimum 3D bounding box enclosing the CAD models is derived from the reference.
In addition, the reference CAD models are used to compute an error metric based on the distances between corresponding keypoints of the reference model and the estimated model in the body coordinate system of the vehicles.
To this end, the RMS error $d_\text{rms}$ of the Euclidean distances between corresponding keypoints is computed for each of the $n_\text{veh}$ vehicles.
This error represents the quality of the estimated shape w.r.t.\ the reference shape for a single vehicle. 
To assess the average quality of shape estimation that is achieved on the whole dataset, the RMSE of $\varepsilon_\text{rms}^\mathcal{K}$ of the distances $d_\text{rms}$ is computed using
\begin{equation}
\varepsilon_\text{rms}^\mathcal{K} = \sqrt{ \frac{1}{n_\text{veh}} \sum_{i=1}^{n_\text{veh}} \left( d_\text{rms}^i \right)^2}.
\end{equation}
To achieve detailed insights into the quality of pose reconstructions, position and orientation estimates are reported in three stages.
The values for $\mathbf{t}_{25}$, $\mathbf{t}_{50}$, and $\mathbf{t}_{75}$ report the percentage of determined vehicle positions whose Euclidean distance $d_\mathbf{t}$ from the reference position is smaller than 0.25\,m, 0.50\,m, and 0.75\,m, respectively.
Similarly, $\theta_5$, $\theta_{10}$, and $\theta_{22.5}$ show the percentage of estimated vehicle orientations whose difference from the reference orientation $d_\theta$ is smaller than $5^\circ$, $10\grad$, and $22.5\grad$, respectively. 
For a joint evaluation of position and orientation, the number of pose estimates that are correct in both, position and orientation, are reported in $\mathbf{t}_{75}+\theta_{5}$ considering the 0.75\,m and $5\grad$ thresholds.
In order to derive a global error metric for all vehicles, a robust measure of the average error is used by reporting the median of the absolute position errors $\varepsilon_\text{Med}^\mathbf{t}$ and the median of the absolute orientation errors $\varepsilon_\text{Med}^\mathbf{\theta}$.
In addition, the median absolute deviation $\sigma_\text{MAD}^{\mathbf{t}/\theta}$ is reported to assess the variability of the errors w.r.t.\ the median with
\begin{equation}
\sigma_\text{MAD}^{\mathbf{t}/\theta} = k \cdot \text{median}\left(\left|d_{\mathbf{t}/\theta}^i - \varepsilon_\text{Med}^{\mathbf{t}/\theta}\right|\right),
\end{equation}
using $k=1.4826$ as a constant factor \citep{Hampel1986}.

\section{Evaluation}
\subsection{Vehicle type prediction}
\begin{figure}[ht]
\centering
\subfloat[Histogram of the \#1 confidence scores obtained by the \textit{vehicle type branch} on the ICSENS data.] {\includegraphics[width=0.75\columnwidth]{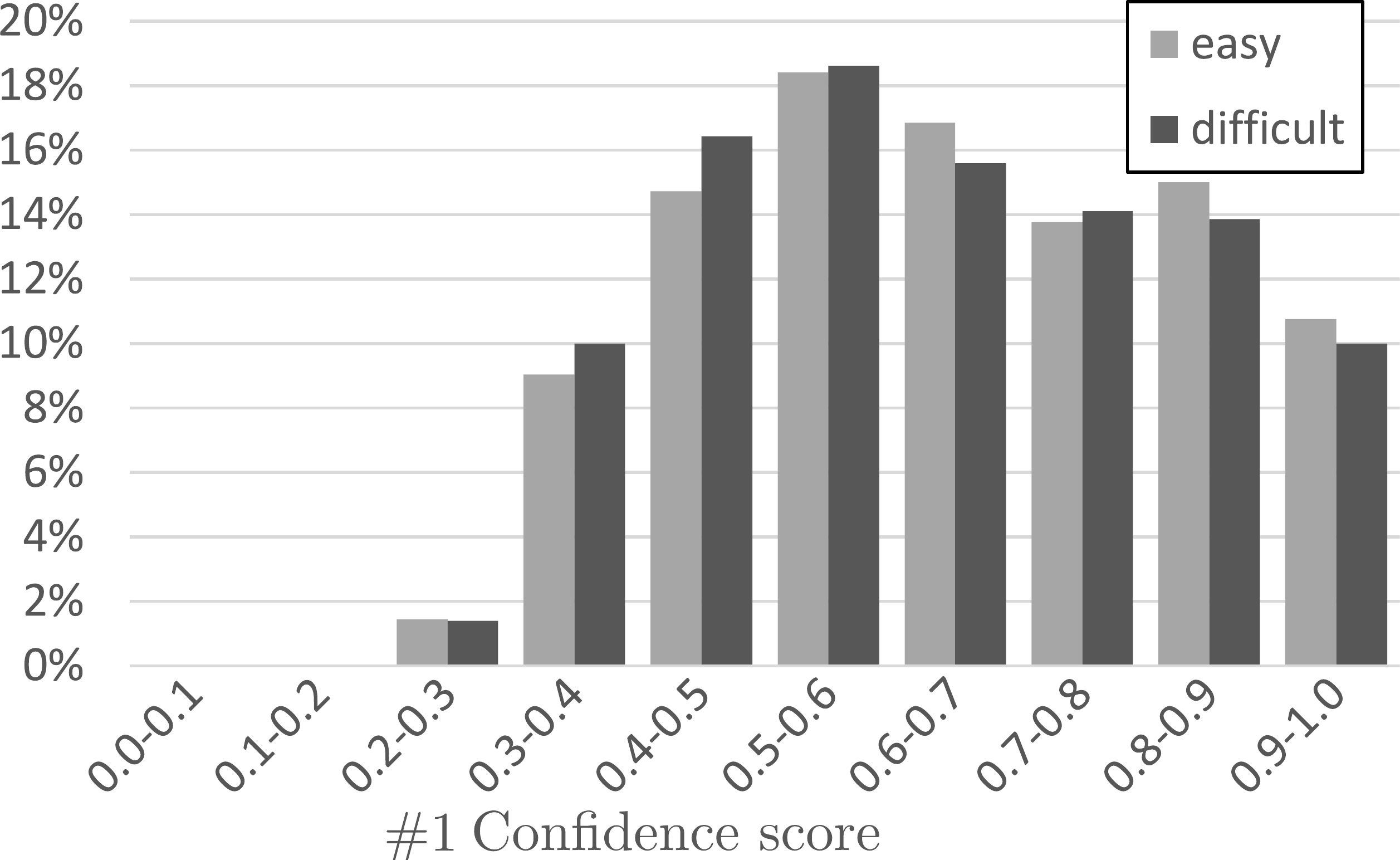}\label{fig:CategoryBranchHist}} \linebreak
\subfloat[Histogram of the ratio of \#2 and \#1 score of the \textit{vehicle type branch} on the ICSENS data..] {\includegraphics[width=0.75\columnwidth]{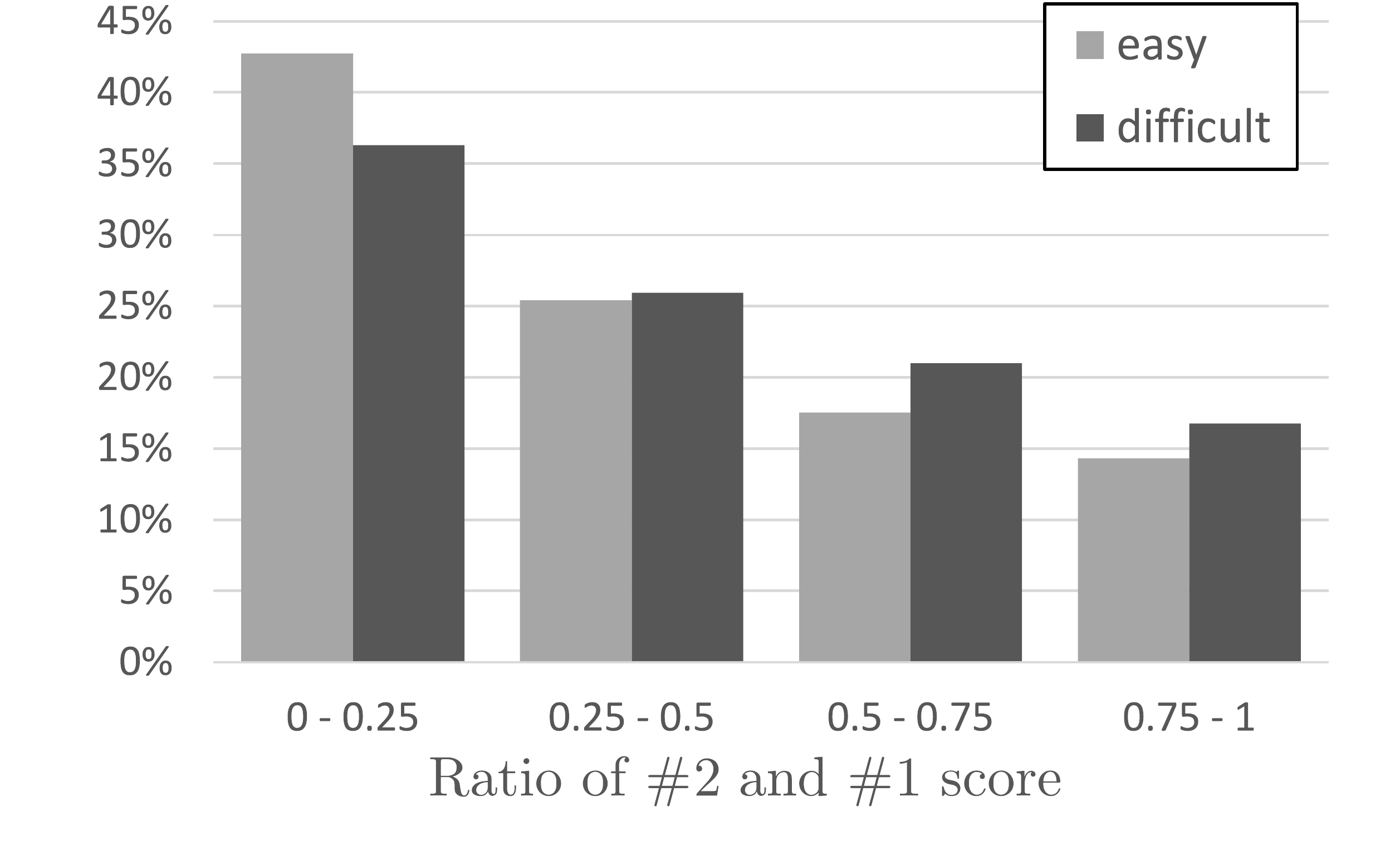}\label{fig:Top1Top2Prop}} 
\caption{Distribution and properties of the classification confidence scores obtained by the \textit{vehicle type branch} on the ICSENS data.}
\label{fig:CategoryBranch} 
\end{figure}

This section provides a study of the performance of the \textit{vehicle type branch} which is proposed in this paper. 
As mentioned in Sec.~\ref{sec:TypeBranch}, the association of a vehicle to one of a set of defined vehicle types can be an ambiguous task in some cases, even for human annotators.
Depending on the vehicle types that are to be distinguished, the transitions between the appearance
of different type classes can be smooth, without clear separation.
As a consequence, for some vehicles, the decision of assigning them to one class or another can be difficult because the definition of the classes is somewhat vague and not stringent.
Transferring these properties to the expected results of a vehicle type classifier leads to the expectation that the partially vague class definitions are reflected by a larger amount of class confusion on the one hand, and by the prediction of confidence scores that lack a distinct maximum on the other hand.
In Tab.~\ref{tab:CategoryClassResults}, the OAs computed from the classification results of the \textit{vehicle type branch} on the ICSENS data set are shown. 
To analyse the effect of potential confusions due to vague class boundaries between specific vehicle types, the Top-1 to Top-3 overall accuracies are reported in the table.
\begin{table}[H]
\centering
\small
\caption{Overall classification results of the \textit{vehicle type branch} on the ICSENS data. Top-1 - Top-3 OAs [\%] are shown.}
		\begin{tabular}{|r |c c c|}\hline		
		 & Top-1 & Top-2 & Top-3 \\ \hline
		easy & 58.0 & 71.6 & 79.8 \\
		difficult & 57.5 & 70.8 & 79.3 \\ \hline
		\end{tabular}		
\label{tab:CategoryClassResults}	
\end{table}
Tab.~\ref{tab:CategoryClassResults} shows that the OA of vehicle types is relatively low. If only the class having the highest score is considered (Top-1 OA), it is at 58.0\% and 57.5\% for the \textit{easy} and \textit{difficult} categories, respectively.
Taking into account the OA achieved for the Top-2 and Top-3 evaluation, which permit confusions between the first two and first three most confidently predicted classes, increases the accuracy by up to 13.6\% and 21.8\%, respectively.
On the one hand, the different data domains used for training and testing can be a potential factor limiting the performance of the classifier on the ICSENS data. 
On the other hand, confusions in the classification results that are caused by the problems described above can potentially cause a significant contingent of incorrect classifications.
To get deeper insights into the behaviour of the \textit{vehicle type branch}, Fig.~\ref{fig:CategoryBranch} shows selected properties of the confidence scores predicted by that branch. 
One indicator for the vagueness of the class definitions to actually be a problem for the classifier is the magnitude of the \#1 confidence score, i.e.\ the score with the largest value among all classes.
A histogram of the \#1 scores obtained by the \textit{vehicle type branch} on the ICSENS data is shown in Fig.~\ref{fig:CategoryBranchHist}. 
Another indicator for the same phenomenon is the ratio between the \#2 and \#1 scores, i.e.\ the ratio between the second largest and the largest confidence scores, lying in the range [0,1].
A larger value for that ratio indicates a higher uncertainty of the classifier about the distinction between the two classes associated to the scores.
A histogram of the ratio between the \#2 and \#1 confidence scores obtained on the ICSENS data is shown in Fig.~\ref{fig:Top1Top2Prop}.
As can be seen in Fig.~\ref{fig:CategoryBranchHist}, only about 10\% of all vehicles obtain a \#1 confidence score of 0.9 or higher for one of the classes.
Instead, the largest amount of more than 18\% of the data achieve \#1 scores between 0.5 and 0.6. 
Interpreting the distribution allows the conclusion that in a considerable amount of cases, the confidence of the classifier to predict the correct class is relatively small. 
As Fig.~\ref{fig:Top1Top2Prop} reveals, this conclusion is further stressed by the relatively large proportion of data (14\% and 16\% for the \textit{easy} and \textit{difficult} levels) for which the ratio of \#2 and \#1 confidence scores is larger than 0.75, i.e.\ the confidence score for the second most probable class is almost as large as the one for the most probable class. Another 18-21\% of the data exhibit a confidence score ratio between 0.5 and 0.75. 

To sum up, the prediction of the vehicle type by the \textit{vehicle type branch} results in comparably low overall Top-1 accuracies.
However, the uncertainties of the prediction are reflected by low confidence scores for the predictions. 
One reason for this behaviour may be the vagueness of the class definition mentioned earlier.
However, the confidence-aware \textit{shape prior} term (Eq.~\ref{eq:ShapePrior}) which makes use of the output of the \textit{vehicle type branch} considers the prediction uncertainties and is therefore able to handle classification errors caused by a potentially indistinct definition of vehicle type assignments.
%
%

\subsection{Shape reconstruction} \label{sec:EvalShape}
This section provides an analysis of the reconstructed vehicle shapes.
In Tab.~\ref{tab:EvalShapeKITTI}, the results achieved on the KITTI data for the shape metrics of the \textit{Base}, \textit{Base+S} and the \textit{Full} models are shown to assess the influence of the \textit{shape prior} term on the shape estimation as well as the final results of the entire probabilistic model. 
Comparing the average absolute errors between the reference and the estimated bounding boxes of the \textit{Base} and the \textit{Base+S} variants, the consideration of the proposed \textit{shape prior} leads to significantly better results for the vehicle length and height, which are improved by up to 7 and 4\,cm, respectively. 
Regarding the error metrics based on the Velodyne points, the consideration of the proposed shape prior leads to an decrease of the median and the average RMSE by 3\,cm and 1\,cm, respectively.
This observation underlines the benefit of the category-aware ASM and \textit{shape prior} presented in this thesis compared to using the commonly applied regularisation of the shape by penalising deviations from the mean ASM shape, as it is done in the \textit{Base} variant. 
The consideration of the \textit{Full} model increases the errors in the estimated length and width of the vehicle extents but gives the distinctly best results for the object height. 
However, regarding the RMSE of the Velodyne points, distinct improvements of the median and average errors of 3 and 2cm, respectively, are achieved by the \textit{Full} model. 
Compared to the base \textit{Base} variant, this is an improvement of 10\% (average) - 20\% (median). 
This observation leads to the conclusion that the \textit{Full} model improves the reconstruction quality of the vehicle parts facing towards the camera, because these are the parts which are covered by the laserscanner observations, and consequently, the increase in the errors of the bounding box extents  results from the estimated vehicle extents in the direction that is invisible to the camera. 
As an additional general observation it can be noted that the errors resulting for the different difficulty levels are of almost the same size, probably due to the applied ASM-based shape prior, which acts as a regulariser on the shape independently from the observability of the vehicles.
\begin{table}[ht]
\centering
\caption{Shape evaluation results on the KITTI data. The dimensions of the reference bounding boxes are compared to the vehicle dimensions resulting from the reconstruction (left part). The median and average RMSE of the velodyne vehicle points to the fitted ASM surface are shown (right part).}
\small
		\begin{tabular}{|c r |c c c | c c|} \hline
		& & \multicolumn{3}{c|}{Average absolute errors [m]} & \multicolumn{2}{c|}{RMSE [m]}\\
		& & \multicolumn{3}{c|}{of bounding box dimensions} & \multicolumn{2}{c|}{of Velodyne points}\\
		& & length & width & height & median & average \\ \hline
		\multirow{3}{0pt}{\rotatebox{90}{easy}} 
		& Base 		& 0.36 &	0.10  &	0.23   									& 0.31	&	0.32\\
		& Base+S 	& \textbf{0.32} &	\textbf{0.08}  &	0.19 	& 0.28	&	0.31\\
		& Full 		& 0.37 &	0.11  &	\textbf{0.12}  					& \textbf{0.25}	&	\textbf{0.29}\\ \hline
		
		\multirow{3}{0pt}{\rotatebox{90}{mod.}} 
		& Base 		& 0.39 &	0.10  &	0.23   									& 0.31	&	0.33\\
		& Base+S 	& \textbf{0.32} &	\textbf{0.09}  &	0.19  & 0.28	&	0.31\\
		& Full 		& 0.37 &	0.11  &	\textbf{0.13}  					& \textbf{0.25}	&	\textbf{0.30}\\ \hline
		
		\multirow{3}{0pt}{\rotatebox{90}{hard}} 
		& Base 		& 0.40 &	0.10  &	0.22   									& 0.31	&	0.35\\
		& Base+S 	& \textbf{0.33} &	\textbf{0.09}  &	0.18 	& 0.28	&	0.33\\
		& Full 		& 0.38 &	0.11  &	\textbf{0.13}  					& \textbf{0.25}	&	\textbf{0.32}\\ \hline
		\end{tabular}		
\label{tab:EvalShapeKITTI}	
\end{table}

Regarding the  errors in object dimensions obtained on the ICSENS data (cf.\ Tab.~\ref{tab:EvalShapeIcsens}), the pattern and magnitude of average absolute errors is comparable to the one obtained on the KITTI data set. The largest discrepancies between reference and estimation occur w.r.t.\ the length of the vehicles, while the vehicle width is estimated with the smallest errors.
Comparing the different tested variants, the smallest errors are obtained by the \textit{Base+S+P+O} setting, in which only the \textit{3D likelihood} under consideration of the state priors is used, which again is an indicator for the benefit of the proposed \textit{shape prior} term.
Comparing the results of the \textit{Full} model to the \textit{Base+S+P+O}, average absolute errors are slightly increased.
This observation is consistent with those made for the KITTI dataset and allows to draw similar conclusions.
The average keypoint error is consistently smallest (24\,cm) for the \textit{Base+S+P+O} variant and slightly increases for the \textit{Full} model.
\begin{table}[H]
\centering
\caption{Shape evaluation results on the ICSENS data. The dimensions of the reference CAD models and the reconstructed ASM are compared. Besides, the keypoint based RMS error is computed from the euclidean distances of corresponding keypoints.}
\small
		\begin{tabular}{|c r |c c c| c|} \hline
		& & \multicolumn{3}{c|}{Average absolute errors [m]} & RMS error \\ 
		& & length & width & height  & $\varepsilon_\text{rms}^\mathcal{K}$ [m] \\ \hline
		\multirow{3}{0pt}{\rotatebox{90}{easy}} 
		& Base+S+P+O 	& \textbf{0.38} &	\textbf{0.09} &	\textbf{0.14}  &		\textbf{0.24}  	\\
		& Base+K+W 		& 0.44 &	0.13 &	0.21  &		0.27 		\\		
		& Full 				& 0.45 &	0.13 &	0.23  &		0.27 		\\ \hline
		
		\multirow{3}{0pt}{\rotatebox{90}{difficult}} 
		& Base+S+P+O 	& \textbf{0.38} &	\textbf{0.10} &	\textbf{0.14} &		\textbf{0.24}		\\
		& Base+K+W 		& 0.43 &	0.12 &	0.20 &		0.26		\\		
		& Full 				& 0.42 &	0.12 &	0.20 &		0.26		\\ \hline
		
		\end{tabular}		
\label{tab:EvalShapeIcsens}	
\end{table}

\subsection{Pose estimation} \label{sec:EvalPose}
The pose estimation results for all variants of the probabilistic model on both, the KITTI and the ICSENS datasets, are presented in Tab.~\ref{tab:PoseResults}, distinguishing between the respective levels of difficulty.
The results will be discussed in the subsequent sections. 
\begin{table*}[ht]
\centering
\small
\caption{Quantitative pose estimation results on the KITTI and ICSENS dataset. The best achieved values for the respective metrics are printed in bold. The arrows indicate if a higher ($\mathbf{\uparrow}$) or a lower ($\mathbf{\downarrow}$) value of the corresponding metric is considered to be better.}
\begin{tabular}{|c r|c c c | c c c | c | c c | c c|}\hline		
& & $\mathbf{t}_{25}$ & $\mathbf{t}_{50}$ & $\mathbf{t}_{75}$
& $\theta_5$ & $\theta_{10}$ & $\theta_{22.5}$ & $\mathbf{t}_{75}+\theta_5$ 
& $\varepsilon^\mathbf{t}_\text{Med}$ & $\sigma^\mathbf{t}_\text{MAD}$ & $\varepsilon^\mathbf{\theta}_\text{Med}$ & $\sigma^\mathbf{\theta}_\text{MAD}$ \\ 
& & \multicolumn{3}{c|}{in [\%] $\mathbf{\uparrow}$}	&	\multicolumn{3}{c|}{in [\%] $\mathbf{\uparrow}$} & 	in [\%] $\mathbf{\uparrow}$ &	\multicolumn{2}{c|}{in [m] $\mathbf{\downarrow}$} &	\multicolumn{2}{c|}{in [$\ \grad$] $\mathbf{\downarrow}$}		\\ \hline \hline

\multicolumn{2}{|c|}{\textbf{KITTI}} & & & & & & & & & & & \\

\multirow{5}{0pt}{\rotatebox{90}{easy}}
& Base 				&39.9 &	71.8 &	87.3 &	57.0 &	68.8 & 	74.5 &	54.1 &	0.31 &	0.25 &	3.9 &	4.5		\\
& Base+S 			&45.0 &	74.9 &	88.8 &	60.4 &	69.5 & 	73.1 &	58.3 &	0.28 &	0.22 &	3.3 &	3.8		\\
& Base+S+P+O 	&\textbf{46.6} &	\textbf{77.1} &	\textbf{92.1} &	77.3 &	95.1 & 	98.6 &	73.0 &	\textbf{0.27} &	\textbf{0.21} &	2.5 &	2.5		\\
& Base+K+W		&40.7 &	73.0 &	88.5 &	84.9 &	95.2 &	97.9 &	77.6 &	0.31 &	0.24 &	1.9 &	1.8		\\
& Full 			&38.6 &	70.3 &	86.6 &	\textbf{90.1} &	\textbf{97.8} & 	\textbf{98.9} &	\textbf{79.8} &	0.33 &	0.26 &	\textbf{1.7} &	\textbf{1.6}		\\ \hline 

\multirow{5}{0pt}{\rotatebox{90}{moderate}}
& Base 				& 36.4 &	67.5 &	82.6 &	52.9 &	64.5 & 	70.2 &	49.7 &	0.34 &	0.27 &	4.5 &	5.4\\
& Base+S 			& 40.9 &	71.0 &	84.3 &	56.0 &	64.9 & 	69.0 &	53.6 &	0.31 &	0.25 &	3.8 &	4.7\\
& Base+S+P+O 	& \textbf{43.0} &	\textbf{74.2} &	\textbf{89.2} &	72.5 &	89.8 & 	95.3 &	68.0 &	\textbf{0.29} &	\textbf{0.22} &	2.7 &	2.7\\
& Base+K+W		& 38.6 &  70.6 &	85.9 &	78.7 &	88.4 &	91.8 &	72.5 &	0.32 &	0.26 &	2.1	& 2.1\\
& Full 			& 37.6 &	69.9 &	86.2 &	\textbf{85.4} &	\textbf{93.5} & 	\textbf{96.2} &	\textbf{76.5} &	0.33 &	0.26 &	\textbf{1.8} &	\textbf{1.7}\\ \hline

\multirow{5}{0pt}{\rotatebox{90}{hard}}
& Base 				&32.5 &	60.8 &	75.2 &	47.7 &	58.0 & 	63.4 &	44.2 &	0.39 &	0.33 &	5.6 &	7.4\\
& Base+S 			&36.3 &	64.0 &	76.9 &	50.4 &	58.4 & 	62.3 &	47.6 &	0.35 &	0.30 &	4.9 &	6.5\\
& Base+S+P+O 	&\textbf{38.9} &	\textbf{68.9} &	\textbf{83.8} &	66.9 &	82.9 & 	89.4 &	61.5 &	\textbf{0.32} &	\textbf{0.26} &	3.1 &	3.2\\
& Base+K+W    &35.4 &	65.0 &	79.8 &	70.3 &	79.6 &	83.2 &	64.7 &	0.36 &	0.30 &	2.4 &	2.7\\
& Full 			&35.0 &	65.7 &	81.8 &	\textbf{78.6} &	\textbf{87.0} & 	\textbf{90.8} &	\textbf{70.1} &	0.36 &	0.29 &	\textbf{2.0} &	\textbf{2.1}\\ \hline \hline

\multicolumn{2}{|c|}{\textbf{ICSENS}} & & & & & & & & & & & \\
\multirow{3}{0pt}{\rotatebox{90}{easy}}
& Base+S+P+O	&\textbf{48.3} &	81.1 &	93.5 &	68.5 &	91.0 &	96.7 &	65.3 &	\textbf{0.26} &	\textbf{0.19} &	3.4 &	3.0\\
& Base+S+K    &38.3 &	76.9 &	90.7 &	69.0 &	88.3 &	94.0 &	65.5 &	0.31 &	0.21 &	3.2 &	2.9\\
&Full       &44.7 &	\textbf{82.5} &	\textbf{93.9} &	\textbf{73.8} &	\textbf{92.2} &	\textbf{97.0} &	\textbf{70.5} &	0.28 &	\textbf{0.19} &	\textbf{2.8} &	\textbf{2.6}\\ \hline

\multirow{3}{0pt}{\rotatebox{90}{diff.}}
& Base+S+P+O	&41.4 &	75.2 &	88.7 &	64.9 &	87.6 &	93.6 &	60.5 &	0.30 &	0.23 &	3.5 &	3.2\\
& Base+S+K    &37.3 &	73.7 &	87.6 &	66.2 &	84.7 &	90.1 &	62.5 &	0.32 &	0.23 &	3.3 &	3.2\\
&Full       &\textbf{43.3} &	\textbf{78.1} &	\textbf{90.5} &	\textbf{71.6} &	\textbf{89.3} &	\textbf{94.1} &	\textbf{67.4} &	\textbf{0.29} &	\textbf{0.21} &	\textbf{2.9} &	\textbf{2.8}\\ \hline

\end{tabular}		
\label{tab:PoseResults}	
\end{table*}

\subsubsection{Results on the KITTI dataset} \ \\
\textbf{Base:} 
In this setting, the ASM is fitted to the triangulated point cloud. As can be seen from Tab.~\ref{tab:PoseResults}, about 40\% of the reconstructed vehicles have a position error of less than 25\,cm and 87.3\% of the vehicles have an error less than 75\,cm in the \textit{easy} category. The median error for the position estimates amounts to 31\,cm.
Because fewer observations are available for vehicles belonging to the \textit{moderate} and \textit{hard} categories, the results for the position estimates are worse, with an median error of up to 39\,cm for the \textit{hard} category. 
The same behaviour can be observed for the orientation estimates, where the median error of $3.9\grad$ achieved for the \textit{easy} level increases to $5.6\grad$ for the \textit{hard} category. 

\textbf{Base+S:}
In this setting the \textit{shape prior} term is added to the model alignment based on the category-aware ASM proposed in this paper. 
As can be noticed in Tab.~\ref{tab:PoseResults}, the number of correct position estimates, especially w.r.t.\ to the finer-grained metrics $\mathbf{t}_{25}$ and $\mathbf{t}_{50}$, is significantly improved by the incorporation of the category-aware shape prior, with improvements of up to 5.1\%.
As has been shown in Sec.~\ref{sec:EvalShape}, the category-aware regularisation of the model shapes results in a better representations of the observed vehicles w.r.t.\ to their dimensions. 
When only considering the 3D likelihood for model alignment, there are errors of shape reconstructions and therefore errors in the estimated vehicle dimensions, caused by the fact that the observed 3D points are only available for a part of the vehicle because no points are observed on the vehicle side facing away from the camera.
Introducing a shape regularisation that is aware of the vehicle category, which in many cases is directly related to the vehicle dimensions, leads to better regularisation constraints on the shape and consequently to enhanced results for the position estimates.
At the same time, the consideration of the \textit{shape prior} also improves the results for orientation estimation w.r.t.\ the $\theta_{5}$ metric.
Conclusively, the category-aware shape prior proposed in this paper leads to an increase in correct position estimates, probably due to a better representation of shape and dimension by the \textit{shape prior}, and to a better quality of the orientation estimations.
This demonstrates the suitability of the type predictions as prior information for the vehicle shape.

\textbf{Base+S+P+O:} 
To assess the full potential of the state priors, the \textit{shape, position}, and \textit{orientation priors} are jointly considered in this variant as regularisers for the state parameters during alignment. 
As can be seen from Tab.~\ref{tab:PoseResults}, throughout all difficulty levels, the consideration of the \textit{position} and \textit{orientation priors} significantly increases the number of correct position and orientation estimates in all metrics and decreases the median errors. 
The analysis of this variant demonstrates the beneficial effect of incorporating the proposed state priors into the approach for model fitting.

\begin{figure*}[ht]
\centering
		\includegraphics[width=0.7\textwidth]{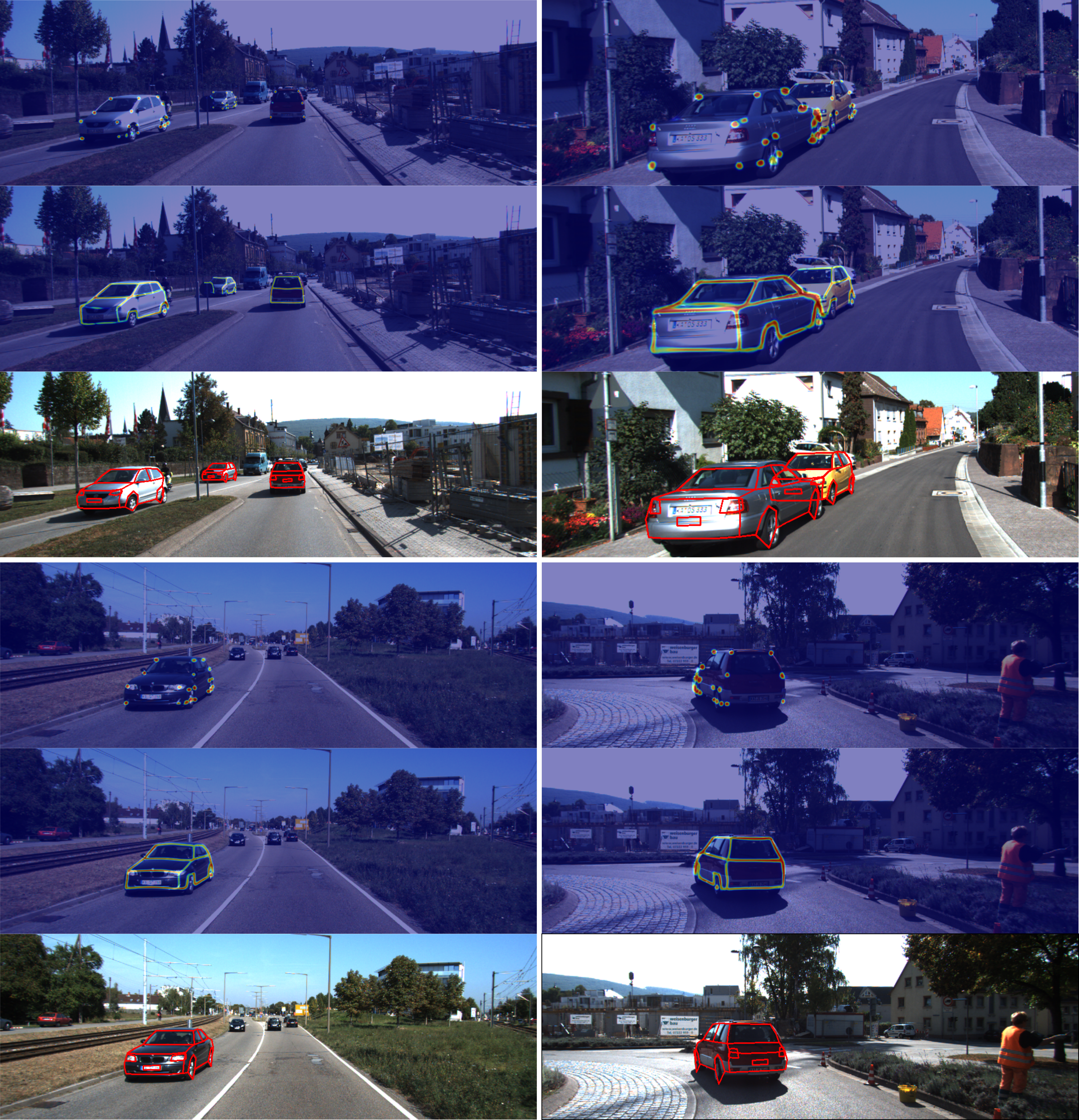}
	\caption{Qualitative results obtained by the \textit{Base+K+W} variant on four images from the KITTI data set. For every image, a triplet consisting of the probability maps for vehicle keypoints (top) and vehicle wireframes (middle) superimposed to the left input image are shown (the cold to warm colour coding represents low to high probabilities). Furthermore, the backprojected wireframes of the reconstructed vehicles are depicted (bottom).}
\label{fig:QualResultsKITTI}
\end{figure*}

\textbf{Base+K+W:}
This setting uses all likelihood terms of the probabilistic model. Compared to the \textit{Base} setting in which only the \textit{3D likelihood} is used, only subtle improvements are achieved for the position estimates but the orientation estimates are enhanced distinctly. In the easy category for instance, the median error for the orientation is decreased from $3.9\grad$ to $1.9\grad$ and the enhancement is even larger for the \textit{moderate} and \textit{hard} levels. It is reasonable to assume that introducing image observations and semantic knowledge via the \textit{keypoint} and \textit{wireframe} likelihoods delivers valuable cues the model fitting. 
This indicates the contribution of the likelihoods to the vehicle reconstruction.
Qualitative results obtained by the \textit{Base+K+W} variant are shown in Fig.~\ref{fig:QualResultsKITTI}, which contains visualisations of the probability maps for the keypoints and wireframes as well as the wireframe of the reconstructed ASM backprojected to the left stereo image. To be able to show the probability
maps for the individual keypoints and wireframe definitions in one image, the heatmaps of all keypoints and wireframes are superimposed and the maximum value among all heatmaps for each pixel is shown.
%
%

\textbf{Full:}
According to Tab.~\ref{tab:PoseResults}, when describing the results of the \textit{Full} probabilistic model compared to the \textit{Base+K+W} and \textit{Base+S+P+O} variants, a distinction has to be made between the results for position and the results for the orientation. 
Counterintuitively, the amount of correct position estimates decreases in the \textit{easy} category when combining the likelihood terms and prior terms in the \textit{Full} model, compared to both, the \textit{Base+K+W} and the \textit{Base+S+P+O} settings.
This decrease is less distinct for the \textit{moderate} and \textit{hard} categories, where the decrease of correct position estimates of the \textit{Full} model w.r.t.\ the \textit{Base+S+P+O} model is smaller, and in fact an increase of correct position estimates using the \textit{Full} model can be observed compared to the \textit{Base+K+W} model. 
The \textit{Base+S+P+O} variant delivers the best results for the position estimates throughout all difficulty levels, with 5.5\% and 8.0\% more correct estimates in the \textit{easy} category for the $\mathbf{t}_{75}$ and $\mathbf{t}_{25}$ metrics, respectively. 
In the \textit{moderate} and \textit{hard} level, the differences are smaller, with 3.0\%/5.4\%, and 2.0\%/3.9\% for the $\mathbf{t}_{75}$/$\mathbf{t}_{25}$ metrics, respectively.

In contrast, regarding the results for the estimated orientations, the \textit{Full} model achieves the highest number of correct estimates throughout all levels of difficulty. 
While the numbers of correct orientation estimates for the $\theta_{22.5}$ metric achieved by the \textit{Base+S+P+O} settings were already fairly high with up to 98.6\% in the \textit{easy} category and up to 89.4\% in the \textit{hard} category, only small improvements of up to 1.4\% are achieved by the \textit{Full} model. The improvements created by the \textit{Full} model compared to the \textit{Base+K+W} and the \textit{Base+S+P+O} models within the finer-grained $\theta_{5}$ metric are up to 12.8\% in the \textit{easy} category and up to 11.7\% in the \textit{hard} category, which is remarkably large. 
The number of vehicle reconstructions that are correct w.r.t.\ both, the $\mathbf{t}_{75}$ and $\theta_{5}$ metrics, are largest for the \textit{Full} model throughout all difficulty levels.
In accordance with the observations described so far, the median errors resulting from the \textit{Full} model are slightly larger compared to the other variants regarding the position, but slightly smaller w.r.t.\ the orientation. 

Comparing the results of the \textit{Full} model to the results reported in \citep{Coenen2019}, the performance of the probabilistic model with and without considering the novel shape prior can be investigated. 
Regarding the $\mathbf{t}_{75}$ metric, the proposed shape prior term improves the results by 7.2\%/5.6\%/4.3\% for the \textit{easy/moderate/hard} categories.
With respect to the orientation, improvements of 3.4\%/3.1\%/2.5\% are achieved for the $\theta_5$ criterion and the different difficulty levels by using the shape prior term proposed in this work.

\subsubsection{Results on the ICSENS dataset}
\begin{figure*}[ht]
\centering
		\includegraphics[width=0.75\textwidth]{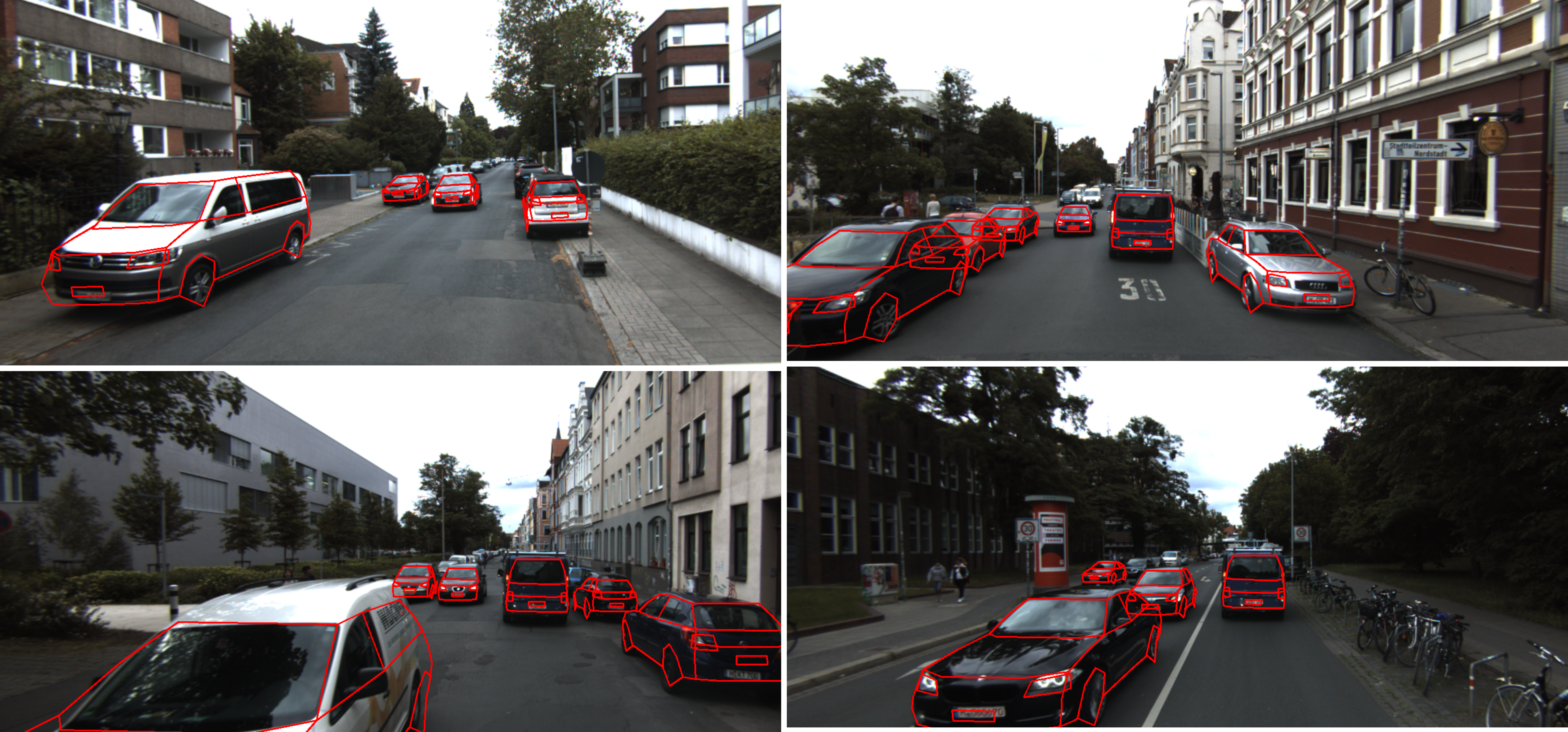}
	\caption{Qualitative results obtained by the \textit{Full} variant on four images from the ICSENS data set. The backprojected wireframes of the reconstructed vehicles are depicted.}
\label{fig:QualResultsICSENS}
\end{figure*}

In addition to the results on the KITTI data set, Tab.~\ref{tab:PoseResults} contains the results for the \textit{Base+K+W}, the \textit{Base\-+S+\-P+O}, as well as the \textit{Full} settings achieved on the ICSENS data.
The benefit of combining all likelihood terms with the state priors in the \textit{Full} formulation on the results for the orientation estimation, which could be observed on the KITTI data, can also be observed in the results achieved for the ICSENS data.
The \textit{Full} model delivers the best results, although the improvement over the other variants is less distinct compared to the results on the KITTI data. 
However, in contrast to the observations made on the KITTI dataset, where the combination of likelihood and prior terms lead to a decrease in the amount of correct position estimates, the \textit{Full} model also improves the position results on the ICSENS data.
An exception is the $\mathbf{t}_{25}$ metric, where the \textit{Base+S+P+O} setting achieves the best results for the \textit{easy} category.
Nevertheless, the overall tendency appearing from Tab.~\ref{tab:PoseResults} attests the beneficial effect of the joint consideration of the proposed likelihoods and state priors.
Qualitative results obtained by the \textit{Full} model on the ICSENS data are shown in Fig.~\ref{fig:QualResultsICSENS}. 

%

%
\subsubsection{Comparison of the results on the KITTI and the ICSENS data sets}
Comparing the results achieved on the two different datasets by the \textit{Full} model, different phenomena can be observed.
Regarding the results for determining the position, it is apparent that the performance on the ICSENS data reveals significant better numbers compared to the performance on the KITTI data.
In the \textit{easy} category, the difference in the amount of correct position estimates ranges from 4.5 - 11.3\% for the different evaluation metrics. 
An explanation for the comparably better results for the position on the ICSENS data can be the larger baseline of the ICSENS stereo rig (85\,cm) compared to the base length used for the KITTI data (54\,cm), and consequently, a comparably lower depth uncertainty. 

Regarding the results obtained for the orientation, the performance of the coarse viewpoint estimation is comparable for both data sets.
However, the achieved numbers for the finer-grained orientation evaluation criteria are distinctly lower for the ICSENS data set, especially for the $\theta_5$ metric, which is 18.4\% and 17.7\% lower in the \textit{easy} and \textit{difficult} levels, respectively, compared to the $\theta_{10}$ criterion. 
As a consequence of the worse orientation estimates, the median errors for the orientation obtained on the ICSENS data are significantly larger compared to the KITTI data.
As the majority of the likelihood and prior terms of the probabilistic model are based on the predictions of the CNN, a potential reason of the decreased performance on the ICSENS data set may be given by the domain gap impacting the performance of the CNN, which is mainly trained on KITTI data and therefore might perform better on data from the same domain.

\subsection{Analysis of limitations and further aspects}
One difficulty in estimating the position is the fact that the position of a vehicle, being represented by the its centre point which is inside of the vehicle, is an entity which is never directly observed in the images, but instead is derived from the reconstructed 3D vehicle shape. 
As a matter of fact, a vehicle is never entirely visible in the image but instead, only one or two of the four vehicle sides are observed, while the remaining parts of the vehicle are averted from the camera and therefore are invisible.
As a consequence, the extent of the vehicle is unobserved and, thus ambiguous. 
Model based approaches as proposed in this paper constrain and derive the full extents of the object using statistically learned shape priors.
As the vehicle position is derived from the reconstructed model, the remaining ambiguities of object extent in the viewing direction cause errors in the position estimates, so that errors are expected to especially occur in the viewing direction of the camera. 
To verify this hypothesis, an analysis of the position estimates, distinguished by errors in lateral (across the viewing direction of the camera) and longitudinal (along the viewing direction of the camera) directions is conducted.
Tab.~\ref{tab:EvalLongLat} contains the numbers of correctly estimated lateral and longitudinal vehicle coordinates, denoted by $\mathbf{t}^\text{lat}$ and $\mathbf{t}^\text{lon}$; for this analysis, a coordinate is considered to be correct if its absolute difference from the reference is smaller than 25, 50 and 75\,cm, respectively.
This table shows the obvious differences between the results for the estimated lateral and longitudinal positions.
\begin{table}[ht]
\centering
\caption{Number of correct longitudinal and lateral position estimates of the \textit{Full} model on the KITTI data.}
\small
		\begin{tabular}{|r |c c c | c c c|} \hline
		& \multicolumn{3}{c|}{Lateral position} & \multicolumn{3}{c|}{Longitudinal position} \\
		$[\%]$ & $\mathbf{t}_{25}^\text{lat}$ & $\mathbf{t}_{50}^\text{lat}$ & $\mathbf{t}_{75}^\text{lat}$ & $\mathbf{t}_{25}^\text{lon}$ & $\mathbf{t}_{50}^\text{lon}$ & $\mathbf{t}_{75}^\text{lon}$ \\ \hline
		easy 			& 88.0 &	97.3 &	98.8 &	43.3 &	73.4 &	88.7 \\
		moderate	& 84.1 &	95.5 &	97.9 &	43.4 &	74.3 &	88.8 \\
		hard			& 79.3 &	91.4 &	95.6 &	42.0 &	71.4 &	85.5 \\ \hline
		\end{tabular}		
\label{tab:EvalLongLat}	
\end{table}
While the number of estimated lateral coordinates that are within 75\,cm of the reference lies between 95.6 and 98.8\% for all difficulty categories, these numbers are approximately 10\% lower for the longitudinal position estimates. This discrepancy becomes even larger considering 25\,cm as the threshold for an estimate to be counted as correct. In the lateral direction, up to 88.0\% of the reconstructed vehicles exhibit a position which is correct within 25\,cm. In the longitudinal direction, only half of this number is achieved. 
The fact that the results for the longitudinal component of the position are worse than those of the lateral one can be attributed to two factors which we believe to interfere with each other. On the one hand, we consider the errors in estimating the vehicle position to be at least partly caused by the ambiguities in determining the spatial extent of the vehicles in the longitudinal direction. On the other hand, this effect can be due to the depth uncertainty, which increases with an increasing distance of the vehicles from the camera.

\begin{figure}[ht]
\centering
		\includegraphics[width=0.3\columnwidth]{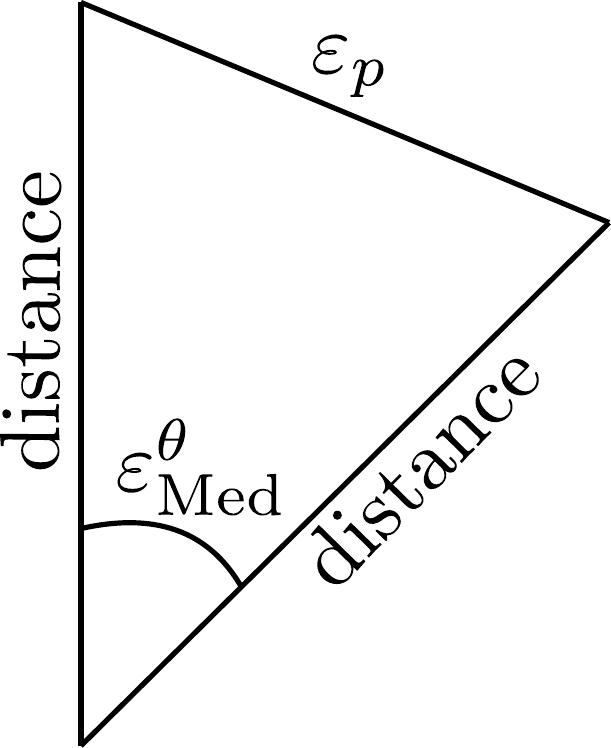}
	\caption{The relation between vehicle distance, orientation error $\varepsilon^\theta_\text{Med}$, and the perpendicular $\varepsilon_p$.}
\label{fig:Perpendicular}
\end{figure}

The influence of the distance of a vehicle from the camera on the quality of the reconstruction can be investigated using Tab.~\ref{tab:ResultsDistance}, which contains the median errors for position and orientation of the \textit{Full} model on the KITTI data, differentiated by the distance of the vehicle to the camera. 
The depth uncertainties $\sigma_x$ of a stereo-reconstructed 3D point in the considered distances, assuming an uncertainty of the disparity $\sigma_\text{disp}$ of 1\,[px], are also shown in Tab.~\ref{tab:ResultsDistance}.
In order to draw a comparison between the quality of the results for the orientation and for the position of the vehicles, the \textit{perpendicular error} $\varepsilon_p$ resulting from the median orientation errors $\varepsilon^\theta_\text{Med}$ in dependency on the distances is computed (cf.\ the Fig.~\ref{fig:Perpendicular}) and can be compared to the median errors of the position $\varepsilon^\mathbf{t}_\text{Med}$ achieved in the respective distances.
The median errors for the orientation achieved on the KITTI data set are used to compute the perpendiculars $\varepsilon_p$ at the corresponding distances in order to compare them to the median errors $\varepsilon^\mathbf{t}_\text{Med}$ for the position achieved on the KITTI data. 
\begin{table}[ht]
\centering
\caption{Median errors for position, orientation, and the perpendicular $\varepsilon_p$ achieved by the \textit{Full} model on the KITTI data differentiated by the vehicle distances.}
\small
		\begin{tabular}{|l r|c c c c|} \hline
		& & \multicolumn{4}{c|}{Vehicle distance} \\ 
		&  & 5\,-\,10\,m & 10\,-\,15\,m & 15\,-\,20\,m & $>$20\,m  \\ \hline
		& $\sigma_x$ [m] & 0.06-0.26 & 0.26-0.58 & 0.58-1.03 & $>$1.03 \\ \hline
		\multirow{3}{0pt}{\rotatebox{90}{easy}} & $\varepsilon^\theta_\text{Med} [\ \grad]$	 & 1.5 & 1.5 & 1.7 & 2.1 \\
		& $\varepsilon_p$ [m] & 0.13 - 0.26 & 0.26 - 0.39 & 0.45 - 0.59 & $>0.92$ \\
		& $\varepsilon^\mathbf{t}_\text{Med}$ [m] & 0.21	&	0.23	&	0.40	&	0.58  \\ \hline	
	  \multirow{3}{0pt}{\rotatebox{90}{moderate}} & $\varepsilon^\theta_\text{Med} [\ \grad]$	 & 1.5 & 1.6 & 1.9 & 2.3  \\
		 & $\varepsilon_p$ [m] & 0.13 - 0.26 & 0.28 - 0.42 & 0.50 - 0.66 & $>1.00$\\
		 & $\varepsilon^\mathbf{t}_\text{Med}$ [m]  & 0.20	&	0.25	&	0.42	&	0.57  \\ \hline		
		 \multirow{3}{0pt}{\rotatebox{90}{hard}} & $\varepsilon^\theta_\text{Med} [\ \grad]$	 & 1.6 & 1.8 & 2.1 & 2.6 \\
		 & $\varepsilon_p$ [m] & 0.14 - 0.28 & 0.31 - 0.47 & 0.55 - 0.73 & $>1.13$ \\
		 & $\varepsilon^\mathbf{t}_\text{Med}$ [m] & 0.21	&	0.28	&	0.44	&	0.60 \\ \hline
		
		\end{tabular}		
\label{tab:ResultsDistance}	 
\end{table}

%
%

Tab.~\ref{tab:ResultsDistance} shows that the distance of the vehicle from the camera strongly affects the ability to correctly estimate the vehicle's position and also affects, although less strikingly, the quality of orientation estimates. 
In the \textit{easy} category, the median error of the position estimates also increases drastically almost by a factor of three from 21\,cm for vehicles in a distance between 5 and 10\,m to 58\,cm for vehicles being more distant than 20\,m. 
The numbers for the \textit{moderate} and \textit{hard} categories are comparable and show the same pattern.
It can be assumed that the increasing depth uncertainty of distant 3D points is responsible for this effect.
While an error of 1\,[px] in the disparity leads to uncertainties $\sigma_x$ of 6-25\,cm for triangulated 3D points in a distance of 5-10\,m, the uncertainty is already 103\,cm for points in a distance of 20\,m. 
As a consequence, and unsurprisingly, the ability to precisely estimate the position is heavily influenced by the expected depth uncertainty.
While an increasing distance of the vehicle also negatively affects the estimation of the orientation, the influence is less distinct compared to the effect on the position estimates.

As can be seen, in a distance larger than 10\,m, the effect of the obtained median errors for the orientation on the perpendicular is larger than the median errors of the position for all difficulty levels. In a distance of 20\,m, $\varepsilon_p$ is almost twice as large as $\varepsilon^\mathbf{t}_\text{Med}$. 
This behaviour can for instance be relevant in the context of applications related to collaborative autonomous driving, in which the determined pose of the vehicles is introduced as \textit{vehicle to vehicle} (V2V) observations for the task of collaborative positioning \citep{CollabPosKnuth}. 

\begin{figure}[H]
\centering
\subfloat[Easy] {\includegraphics[width=0.9\columnwidth]{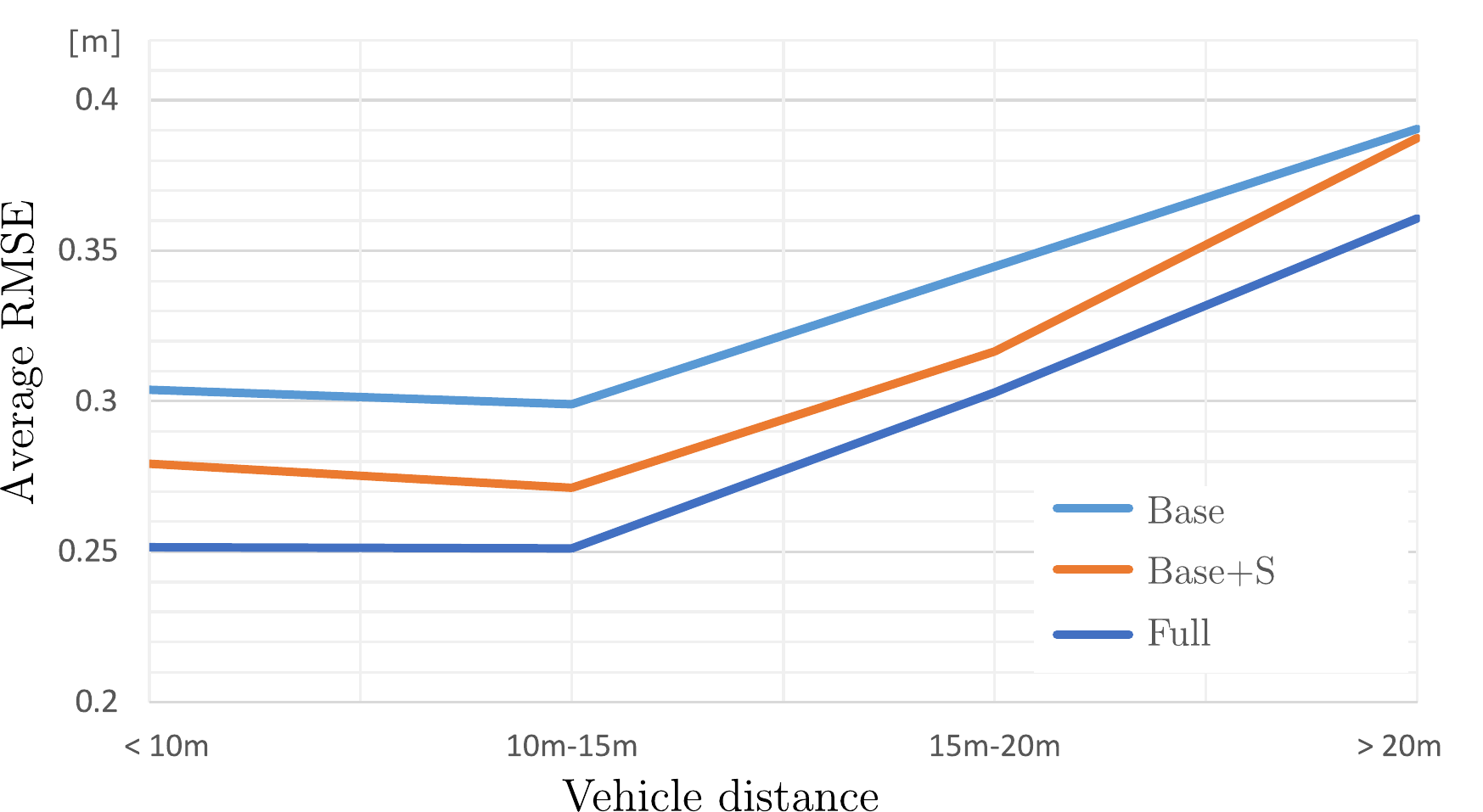}\label{fig:VeloEasy}} \linebreak
\subfloat[Moderate] {\includegraphics[width=0.9\columnwidth]{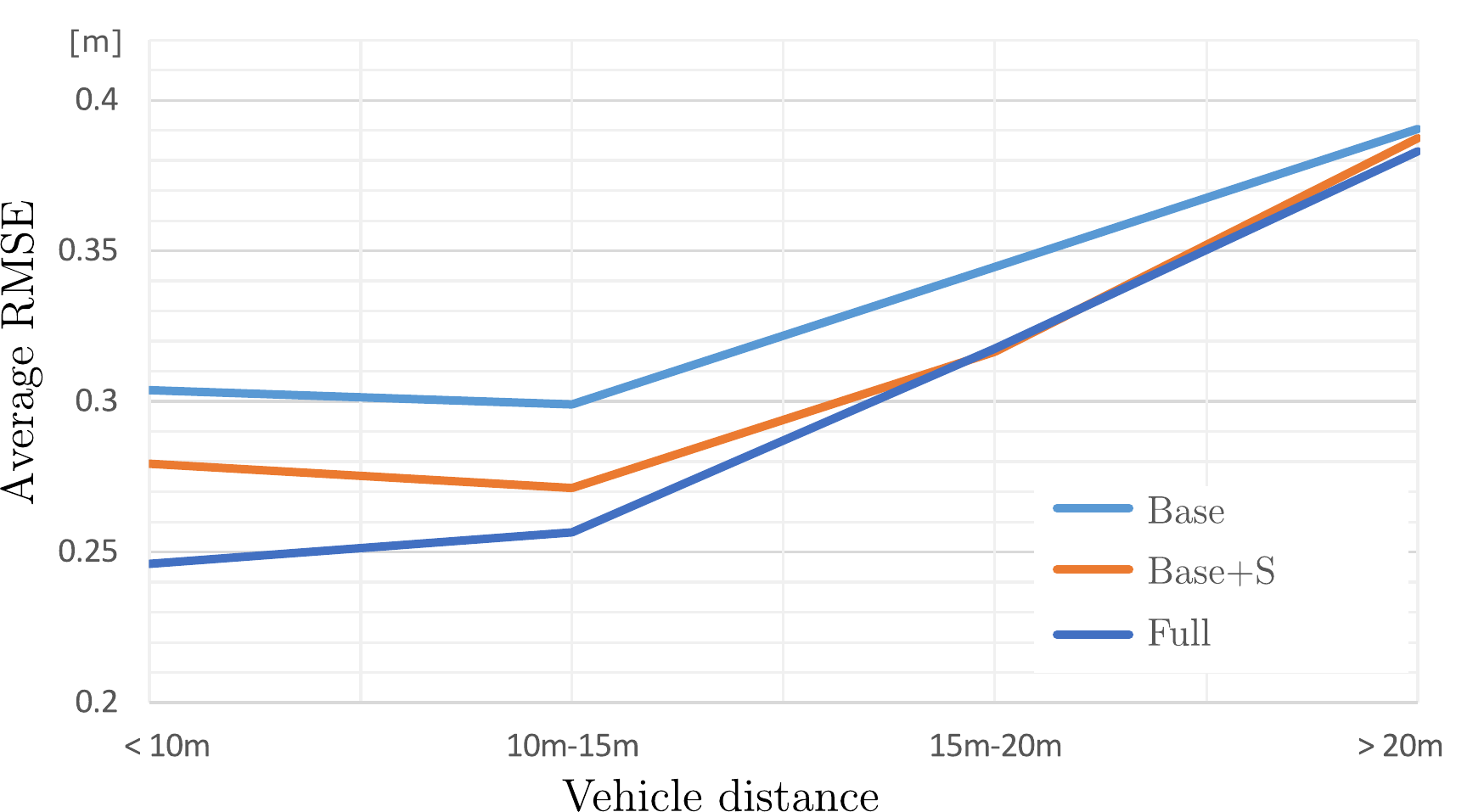}\label{fig:VeloModerate}} \linebreak
\subfloat[Hard] {\includegraphics[width=0.9\columnwidth]{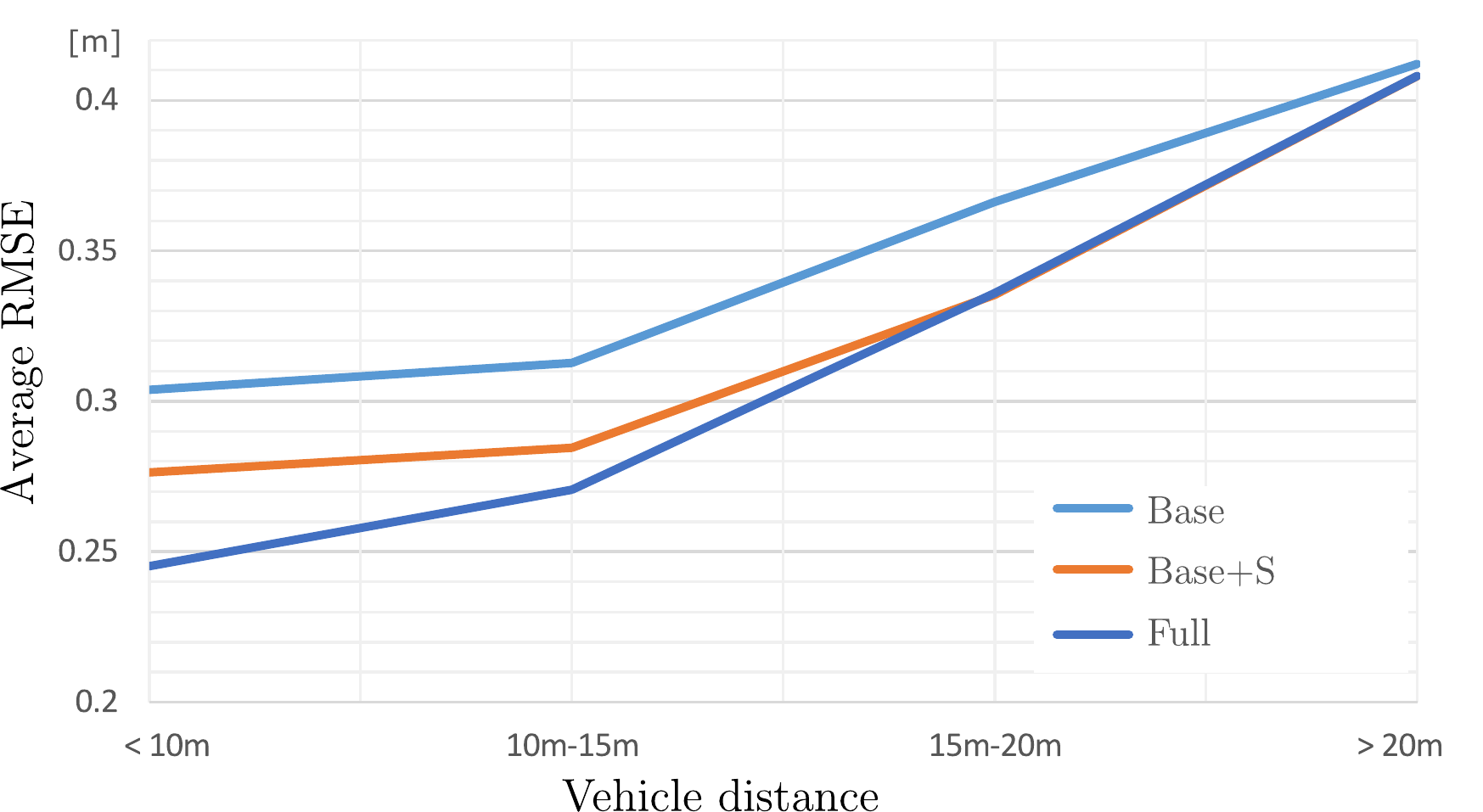}\label{fig:VeloHard}} 
\caption{Shape evaluation results on the KITTI data based on the velodyne laserscanner points. The average RMSE of the laserscanner vehicle clouds to the ASM surface is shown as a function of the vehicle distance to the camera for the different variants of the probabilistic model.}
\label{fig:VeloEval} 
\end{figure}

In Fig.\ref{fig:VeloEval}, the average RMSE of the Velodyne points and the reconstructed ASM surface are shown as a function of the vehicle distance to the camera.
%
%
It can be seen that the RMSE significantly increases as the vehicle distance becomes larger than 15m. 
What becomes visible, too, is that the consideration of the proposed shape prior in the \textit{Base+S} variant and the consideration of the \textit{Full} model lead to the largest improvements for vehicles close to the camera (up to 15\,m); 
the improvements become smaller with an increasing  distance of the vehicles from the camera. 
We assume this to be the result of the increasing depth errors of the reconstructed points which form the basis of 3D reconstruction (cf.~Tab.~\ref{tab:ResultsDistance}): for points at a large distance from the camera, the deviations of the reconstructed models from the laser point cloud are dominated by these depth errors.

\subsection{Comparison to related methods}
In order to compare the performance of the proposed method to the performance of related methods, the test data set and evaluation metrics of the official KITTI benchmark\footnote{http://www.cvlibs.net/datasets/kitti} can be used.
However, the focus of that benchmark lies on object detection.
As a consequence, metrics to assess the quality of pose estimation are coupled with the performance of detection; cf. \citep{KITTI} for a detailed description of the error metrics.
To assess the orientation accuracy, the official metric of the KITTI benchmark is the Average Orientation Similarity (AOS), which multiplies the Average Precision (AP) of the detector with the average cosine distance similarity for the orientation.
Because the focus of this work lies on the pose estimation rather than the detection of vehicles, we make use of the \textit{Orientation Score} (OS) metric which was proposed by \citet{Mousavian2017}.
OS is a metric which factors out the 2D detector performance by computing the ratio between AOS over AP and thus, is unaffected by the detector's performance and can be used to assess the quality of the estimated orientation.
Tab.~\ref{tab:Comparison} shows the OS scores achieved by our \textit{Full} approach and by related state-of-the-art approaches.
\begin{table}[H]
\centering
\caption{Comparison to related methods based on the performance of the KITTI test set. The \textit{orientation score} (OS) allows a mere comparison of the orientation estimates without the impact of the detection performance.}
\small
		\begin{tabular}{| r | c|  c|  c|} \hline
		OS scores & easy & moderate & hard\\ \hline
		\citet{DPMview}				& 96.28	& 95.29	& 95.06	\\
		\citet{3DOP} 					& 98.28 & 97.13 & 97.73 \\
		\citet{Chen2016}			& 98.57 & 97.69 & 97.31 \\
		\citet{Xiang2017}			& 99.84 & 99.52 & 99.25 \\ 
		\citet{Ku2019}				& 99.63	& 98.81	& 98.50 \\
		\citet{Manhardt2019} 	& 98.38	& 97.12	& 96.45	\\ \hline
		 Ours (Full)					& 98.77 & 97.65 & 96.75 \\ \hline
		\end{tabular}		
\label{tab:Comparison}	
\end{table}
As can be seen from the table, our method is on par with or outperforms most of the shown related methods in the \textit{easy} and \textit{moderate} categories.  
Regarding the \textit{hard} category, our methods reveals a larger decrease of the OS compared to the related methods. 
Throughout all difficulty levels, the best results for OS are achieved by the methods presented by \citet{Xiang2017} and \citet{Ku2019}. 
However, it has to be noted that the different CNN architectures for pose estimation which are presented in these papers were trained using the entire KITTI training data set of almost 8000 images from the same domain, while the approach presented in this paper was trained on  260 images from the KITTI training set only.

\begin{table*}[ht]
\centering
\caption{Comparison of the proposed method to the results achieved by \citet{Mousavian2017} based on the evaluation metrics proposed in this paper.}
\small
		\begin{tabular}{|c r |c c c| c c c| c|} \hline
		 & in [\%] & $\mathbf{t}_{25}$ & $\mathbf{t}_{50}$ & $\mathbf{t}_{75}$ & $\theta_{5}$ & $\theta_{10}$ & $\theta_{22.5}$ & $\mathbf{t}_{75} + \theta_{5}$ \\ \hline
		\multirow{2}{0pt}{\rotatebox{90}{easy}}
		&\citet{Mousavian2017} & 19.5 & 38.4 & 55.1 & 87.1 & 94.9 & \textbf{98.9} & 54.8 \\
		&Ours (Full) & \textbf{42.7} & \textbf{75.2} & \textbf{89.5} & \textbf{88.7} & \textbf{97.0} & 98.5 & \textbf{88.5} \\ \hline
		\multirow{2}{0pt}{\rotatebox{90}{mod.}}
		&\citet{Mousavian2017} & 16.8 & 34.0 & 49.2 & 78.7 & 90.1 & \textbf{96.1} & 48.1  \\
		&Ours (Full) 					& \textbf{39.3} & \textbf{72.4} & \textbf{87.8} & \textbf{83.5} & \textbf{92.3} & 95.6 & \textbf{85.8}  \\ \hline
		\multirow{2}{0pt}{\rotatebox{90}{hard}}
		&\citet{Mousavian2017} & 13.9 & 29.1 & 42.9 & 68.6 & 80.2 & 88.1 & 40.4  \\
		&Ours (Full) 					& \textbf{35.5} & \textbf{66.9} & \textbf{81.9} & \textbf{75.5} & \textbf{84.7} & \textbf{89.0} & \textbf{78.6}  \\ \hline
		\end{tabular}		
\label{tab:ComparisonMousavian}	
\end{table*}

In order to compare the evaluation metrics developed in the context of this paper to results achieved by other state-of-the-art approaches, we make use of the detection and pose estimation results of \citep{Mousavian2017}, which are provided by the authors for a subset of 3799 images from the KITTI training data. 
Tab.~\ref{tab:ComparisonMousavian} contains the comparison of results of the proposed \textit{Full} model and the results from \citet{Mousavian2017} on the same set of images. The images that were used for training in this work were excluded from the evaluation of both, the results of \citet{Mousavian2017} and ours. 

Our method significantly outperforms \citet{Mousavian2017} in terms of the number of correct position estimates, especially for the fine evaluation metric $\mathbf{t}_{25}$, by a factor of up to 2.5. However, for a fair comparison it has to be noted that in \citep{Mousavian2017} no stereo information is used. Regarding the orientation estimates, while the results for the coarse level of $\theta_{22.5}$ are similar, the probabilistic model of this work leads to significantly better results for the finer evaluation metrics, particularly for the \textit{moderate} and \textit{hard} categories. As a consequence, the number of vehicle reconstructions that are considered as correct in both, position and orienation, obtained by our approach is considerably larger.

\section{Conclusion}
In this paper, we proposed a subcategory-aware shape prior for vehicles.
Together with a CNN based prediction of the vehicle type, the novel shape prior is incorporated into an extensive probabilistic model for vehicle reconstruction. 
The results on two real-world datasets have shown the benefit of the proposed shape prior w.r.t. both, the quality of vehicle shape reconstructions and of vehicle pose estimates. 
A comparison of our results to those of related state-of-the-art methods has shown that the proposed approach performs on par or better, confirming the suitability of the developed shape prior and probabilistic model for vehicle reconstruction. 
While the proposed method provides an approach for vehicle reconstruction based on one individual stereo pair, i.e. observations from one epoch in time, an extension of the probabilistic model to a sequence of images acquired at subsequent epochs and the tracking of vehicles would enable the incorporation of multiple observations of the same object.
Suitable constraints, e.g. on plausible changes in the position and orientation between two time
steps, can for instance be enforced by proper motion models.
Besides, the reconstruction of each vehicle is treated individually in this paper.
Global constraints that e.g.\ prevent vehicle reconstructions to coincide, or contextual relations between vehicles are not exploited so far but provide opportunities for further improvement.
Another future direction of research could be related to the improvement of the CNN training procedure to allow for end-to-end training and to improve the results of the vehicle type prediction branch. 
Finally, further experiments can highlight whether the introduction of relative weights to the components of the probabilistic model could further improve the results. 

\section*{Acknowledgements}
This work was supported by the German Research Foundation (DFG) as part of the Research Training Group i.c.sens [GRK2159]. 



 \bibliographystyle{elsarticle-harv} 



\bibliography{Literatur}
\end{document}